\documentclass{article} 
\usepackage{iclr2025_conference,times}

\usepackage{amsmath,amsfonts,bm}

\def\eqref#1{equation~\ref{#1}}

\def\1{\bm{1}}

\DeclareMathAlphabet{\mathsfit}{\encodingdefault}{\sfdefault}{m}{sl}
\SetMathAlphabet{\mathsfit}{bold}{\encodingdefault}{\sfdefault}{bx}{n}

\usepackage{hyperref}
\usepackage{url}
\usepackage[utf8]{inputenc} 
\usepackage[T1]{fontenc}    
\usepackage{hyperref}       
\usepackage{url}            
\usepackage{booktabs}       
\usepackage{amsfonts}       
\usepackage{nicefrac}       
\usepackage{microtype}      
\usepackage{xcolor}         
\usepackage{wrapfig}
\usepackage{adjustbox}
\usepackage{multirow}
\usepackage{caption}
\usepackage{comment}
\usepackage{bm}
\usepackage{cellspace}
\usepackage{tabularx}
\usepackage{makecell} 
\usepackage[font=small]{caption}
\usepackage[font=footnotesize]{subcaption}
\title{Faceshot: Bring any Character into Life}

\author{{Junyao Gao}$^{1}$\thanks{Work done during an internship in Shanghai AI Laboratory. \textsuperscript{\ddag} corresponding authors.}, {Yanan Sun}$^{2}$\textsuperscript{\ddag}, {Fei Shen}$^{3}$, {Xin Jiang}$^{3}$, {Zhening Xing}$^{2}$\\
\textbf{{Kai Chen}}$^{2}$\textsuperscript{\ddag}\textbf{,} \textbf{{Cairong Zhao}}$^{1}$\footnotemark[2]\textsuperscript{\ddag}\\
$^{1}$Tongji University, $^{2}$Shanghai AI Laboratory, $^{3}$Nanjing University of Science and Technology.\\
{\tt\small \texttt{\{junyaogao,zhaocairong\}@tongji.edu.cn, \{feishen,xinjiang\}@njust.edu.cn},}\\
{\tt\small \texttt{\{sunyanan,xingzhening,chenkai\}@pjlab.org.cn.}}
}

\iclrfinalcopy
\begin{document}

\maketitle

\begin{figure}[h]
\vspace{-4mm}
	\centering
        \includegraphics[width=1.0\linewidth]{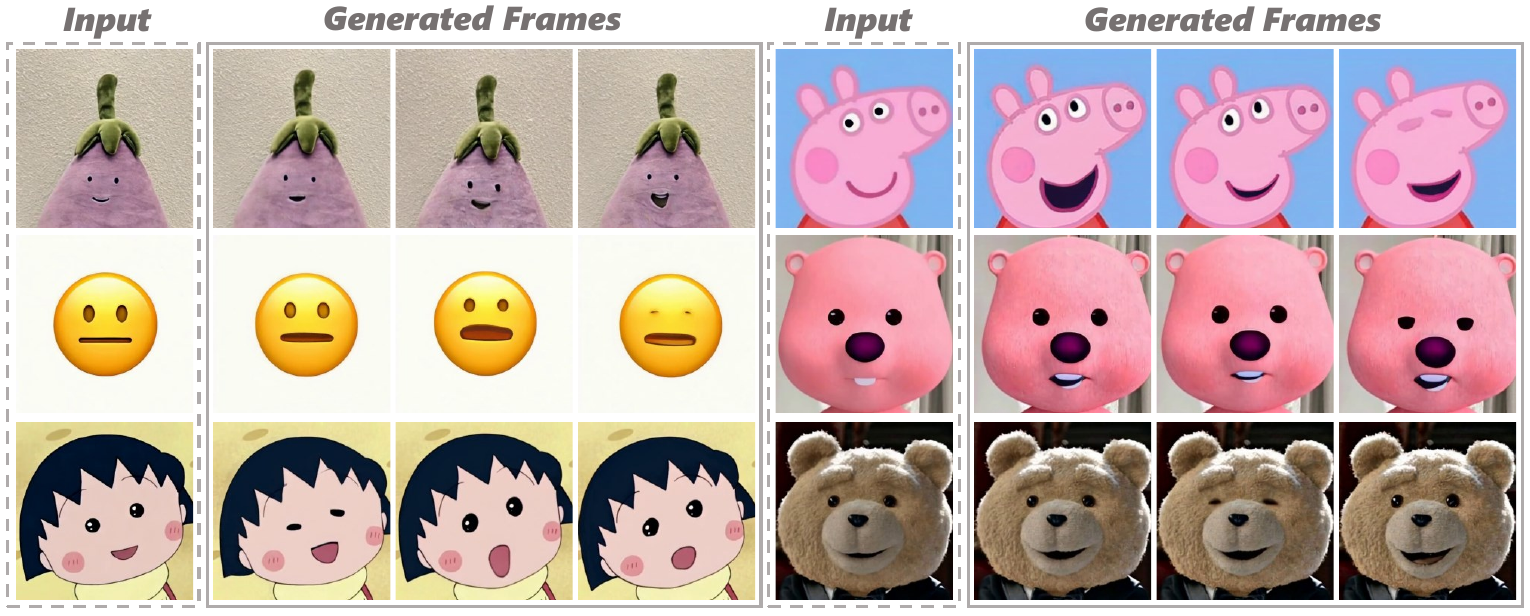}
	\caption{Visualization results of our \textbf{FaceShot}. Given any character and any driven video, FaceShot effectively captures subtle facial expressions and generates stable animations for each character. Especially for non-human characters, such as emojis and toys, FaceShot demonstrates remarkable animation capabilities.}
	\label{fig:teaser}
\end{figure}

\begin{abstract}
In this paper, we present \textbf{FaceShot}, a novel training-free portrait animation framework designed to bring any character into life from any driven video without fine-tuning or retraining.
We achieve this by offering precise and robust reposed landmark sequences from an appearance-guided landmark matching module and a coordinate-based landmark retargeting module.
Together, these components harness the robust semantic correspondences of latent diffusion models to produce facial motion sequence across a wide range of character types.
After that, we input the landmark sequences into a pre-trained landmark-driven animation model to generate animated video.
With this powerful generalization capability, FaceShot can significantly extend the application of portrait animation by breaking the limitation of realistic portrait landmark detection for any stylized character and driven video.
Also, FaceShot is compatible with any landmark-driven animation model, significantly improving overall performance.
Extensive experiments on our newly constructed character benchmark CharacBench confirm that FaceShot consistently surpasses state-of-the-art (SOTA) approaches across any character domain.
More results are available at our project website \url{https://faceshot2024.github.io/faceshot/}.
\end{abstract}

\section{Introduction}
"I wish my toys could talk"-many people make this wish on birthdays or at Christmas, hoping for the companionship from their ``imaginary friends''.
Achieving this usually requires a bit of ``magic'', as seen with the talking teddy bear in \textit{Ted}\footnote{\url{https://en.wikipedia.org/wiki/Ted_(film)}} or the three chipmunks in \textit{Alvin and The Chipmunks}\footnote{\url{https://en.wikipedia.org/wiki/Alvin_and_the_Chipmunks_(film)}}.
Behind these productions, making this ``magic'' a reality often requires specialized equipment and significant manual effort for character modeling and rigging.
In this work, to bring any character into life for every person, we propose a novel, training-free portrait animation framework.
As shown in Figure \ref{fig:teaser}, even for emojis and toys, which have totally different facial appearances from humans, our proposed framework demonstrates remarkable performance in making these characters alive.

\begin{wrapfigure}{r}{0.7\linewidth}
		\centering
  \vspace{-4mm}
		\includegraphics[width=\linewidth]{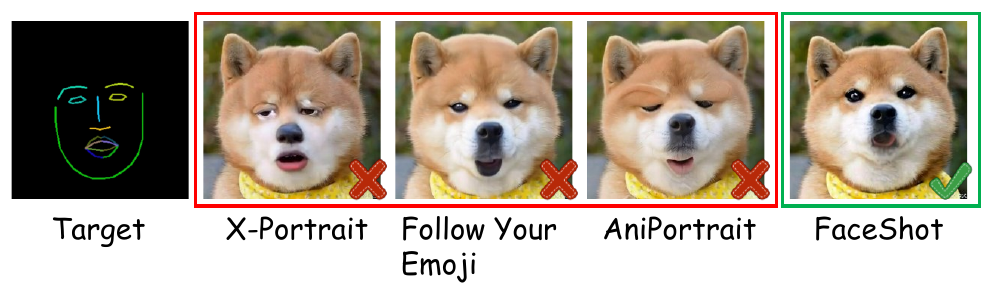}
        \vspace{-4mm}
\caption{Visual results generated from current portrait animation methods and our FaceShot. Previous methods apparently retain the target human's appearance. In contrast, the result of FaceShot both aligns the dog's facial features and captures the target human's expression.}
\label{fig:intro}
\end{wrapfigure}

Portrait animation \citep{guo2024liveportrait, ma2024follow, xie2024x, yang2024megactor, wei2024aniportrait, niu2024mofa, wang2021one, zeng2023face} has demonstrated impressive results with the recent advancements in generative models, such as Generative Adversarial Networks (GANs) \citep{gan, donahue2016adversarial, odena2017conditional, radford2015unsupervised} and diffusion models \citep{rombach2022high, nichol2021glide, saharia2022photorealistic, ho2020denoising,song2020denoising}.
However, these methods depend on facial landmark recognition, and their performance are constrained by the generalization capability of facial landmark detection models \citep{zhou2023star, yang2023unipose}. 
For non-human characters, such as emojis, animals and toys, which often exhibit significantly different facial features compared to human portraits, always resulting in landmark recognition failures due to the supervised training paradigm and limited datasets.
As shown in Figure \ref{fig:intro}, the unaligned target facial features for non-human character lead to disrupted animation results; the animation models even generate a human mouth in the wrong position for a dog.
In addition, \cite{xie2024x, yang2024megactor} indicate that these methods cannot control subtle facial motions, leading to inconsistent animation results.

To address the above limitations, we propose \textbf{FaceShot}, a novel portrait animation framework capable of animating any character from any driven video without the need for training.
As demonstrated in Figure \ref{fig:teaser}, FaceShot produces vivid and stable animations for various characters, particularly for non-human characters. 
This is achieved through three key components: (1) the appearance-guided landmark matching module, (2) the coordinate-based landmark retargeting module, and (3) a landmark-driven animation model.

In our landmark matching module, we inject appearance priors into diffusion features and leverage their strong semantic correspondences to match the landmarks. 
For the second component, we introduce a theoretical algorithm to capture subtle facial motions between frames and generate the landmark sequence aligned with the driven video.
Finally, for the third component, we input the landmark sequence of the reference character into a pre-trained landmark-driven animation model to animate the character.
As shown in Figure \ref{fig:intro}, FaceShot provides the reasonable animation result by offering the precise landmarks of non-human character.
Furthermore, FaceShot is compatible with any landmark-driven portrait animation model as a plugin, improving their performance on non-human characters, with experimental analysis in Section \ref{sec:abla}.

Moreover, to address the absence of a benchmark for character animation, we establish CharacBench that contains 46 diverse characters.
Qualitative and quantitative evaluations on CharacBench demonstrate that FaceShot excels in animating characters, especially in non-human characters, outperforming existing portrait animation methods. 
Additionally, ablation studies validate the effectiveness and superiority of our framework, providing valuable insights for the community. 
Furthermore, we provide the animation results of FaceShot from non-human driven videos, bringing a potential solution to the community for open-domain portrait animation. 

The main contributions of this paper are as follows:
\begin{itemize}
     \item We propose FaceShot, a novel portrait animation framework capable of animating any character from any driven video without the need for training.
     \item FaceShot generates precise reposed landmark sequences for any character and any driven video, bringing a potential solution to the community for open-domain portrait animation.
     \item FaceShot can be seamlessly integrated as a plugin with any landmark-driven animation model, further improving its performance.
     \item We establish CharacBench, a benchmark with diverse characters for comprehensive evaluation. Experiments on CharacBench show that FaceShot outperforms SOTA approaches.
\end{itemize}

\section{Related Work}

\noindent{\textbf{Portrait Animation.}}
Early portrait animation methods primarily relied on GANs~\citep{gan} to generate portrait animation through warping and rendering techniques~\citep{drobyshev2022megaportraits, siarohin2019first, hong2022depth, wang2021one, zhao2022thin}.
Recent advancements in latent diffusion models (LDMs)~\citep{rombach2022high, ramesh2022hierarchical, shen2023advancing, gao2024styleshot, shen2024boosting,wang2024ensembling,shen2024imagdressing,li2024towards,shen2024imagpose} have improved the quality and efficiency of image generation. 
Building on this, some methods \citep{xie2024x, yang2024megactor} learn the identity-free expression by constructing paired data, demonstrating impressive performance.
However, current data collection pipelines face challenges in constructing paired data for diverse domains, particularly for non-human characters, which limits the generalization ability of these methods.
Additionally, other approaches \citep{wei2024aniportrait, niu2024mofa, ma2024follow, shen2025long} tend to utilize highly scalable conditions, such as facial landmarks, to enhance motion control.
Naturally, these methods depend on the facial landmark recognition, which also restricts their applicability to non-human characters.
To break these limitations and bring any character into life, we focus on providing precise reposed landmark sequences within landmark-driven portrait animation in this work.

\noindent{\textbf{Facial Landmark Detection}}
Facial landmark detection aims to detect key points in given face.
Traditional methods \citep{cootes2000introduction, cootes2001active, dollar2010cascaded, kowalski2017deep} often construct a shape model for each key point and perform iterative searches to match the landmarks.
With the development of deep networks, some methods \citep{sun2013deep, zhou2013extensive, wu2018look, wu2017facial} select a series of coarse to fine cascaded networks to perform direct regression on the landmarks.
Another trend, \cite{huang2020propagationnet, zhou2023star, merget2018robust, kumar2020luvli} predict the heatmap of each point for indirect regression, improving the accuracy of landmark detection.
Recently, \cite{yang2023unipose, xu2022pose, li2024cascaded} collect larger datasets and train larger models for more generalize landmark detection.
However, due to the supervised training paradigm and limited dataset, these methods are difficult to perfectly detect the landmark of non-human characters.
In our paper, we turn to a training-free landmark matching module through the strong semantic correspondence and the generalization in diffusion features \citep{tang2023emergent, hedlin2024unsupervised, luo2024diffusion}, aiming to provide precise landmarks for non-human characters.

\noindent{\textbf{Image to Video Generation}}
Image to video (I2V) generation has gained significant attention in recent years due to its potential in various applications, such as image animation \citep{dai2023animateanything, gong2024atomovideo, ni2023conditional, guo2023animatediff} and video synthesis \citep{blattmann2023align, wang2024videocomposer, ruan2023mm}.
Since diffusion models demonstrate the powerful image generation capabilities, \cite{zhang2024pia, shi2024motion, xing2023dynamicrafter, ma2024follow} achieve image animation by inserting temporal layers into a pre-trained 2D UNet and fine-tuning it with video data.
Furthermore, some methods \citep{zhang2023i2vgen, blattmann2023stable}  have constructed their own I2V models and performed full training with large amounts of high-quality data, demonstrating strong competitiveness.
In our work, we utilize MOFA-Video \citep{niu2024mofa} as our base animation model.

\section{Method} \label{headings}
\begin{figure}[t]
    \centering
    \vspace{-4mm}
    \includegraphics[width=1.0\linewidth]{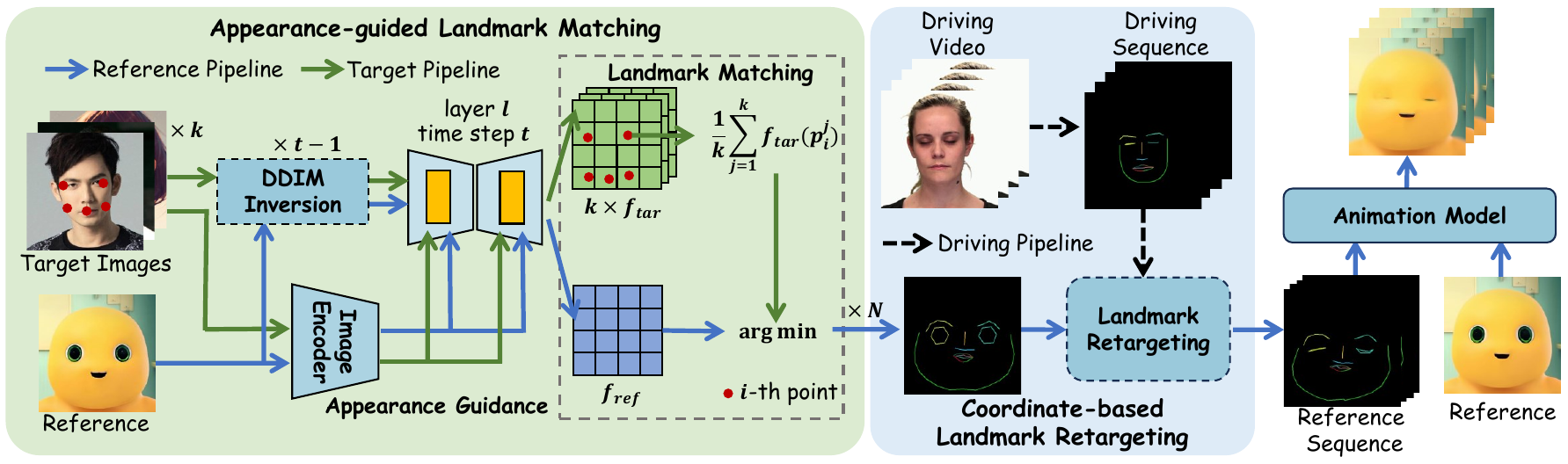}
    \caption{
    The FaceShot framework first generates precise facial landmarks for the reference character with appearance guidance. Next, a coordinate-based landmark retargeting module generates the landmark sequence based on driving video. Finally, this sequence is fed into an animation model to animate character.}
    \vspace{-4mm}
    \label{fig:framework}
\end{figure}

The framework of FaceShot is depicted in Figure~\ref{fig:framework}. We first introduce the foundational concepts of diffusion models in Section \ref{sec:pre}. 
We then explain the three key components of our framework in detail in Section \ref{sec:method}: appearance-guided landmark matching, coordinate-based landmark retargeting, and character animation model.

\subsection{preliminary}
\label{sec:pre}
In FaceShot, we utilize Stable Diffusion (SD) \citep{rombach2022high} as the base model for landmark matching, which consists of a Variational Auto-Encoder (VAE)~\citep{vae}, a CLIP text encoder~\citep{clip}, and a denoising U-Net~\citep{ronneberger2015u}. 
Compared to pixel-based diffusion models, SD uses the VAE encoder $\mathcal{E}$ to encode the input image $\mathbf{x}$ into a latent representation $\mathbf{z} = \mathcal{E}(\mathbf{x})$. 
The VAE decoder $\mathcal{D}$ then reconstructs the image by decoding the latent representation: $\mathbf{x} = \mathcal{D}(\mathbf{z})$.

To train the denoising U-Net ${\epsilon_\theta}$, the objective typically minimizes the Mean Square Error (MSE) loss $\mathcal{L}$ at each time step $t$, as follows:
\begin{equation}
    \mathcal{L} = \mathbb{E}_{\mathbf{z}^t,\epsilon \sim \mathcal{N}(\mathbf{0}, \mathbf{I}), \mathbf{c}, t} \Vert \epsilon_{\theta}(\mathbf{z}^t, \mathbf{c}, t) - \epsilon^{t} \Vert^{2},
\end{equation}
where $\mathbf{z}^t = \sqrt{\Bar{\alpha}^t}\mathbf{z}^0 + \sqrt{1 - \Bar{\alpha}^t}\epsilon^t$ is the noisy latent at time step $t$, $\Bar{\alpha}^t:=\prod_{s=1}^t \alpha^s$ and $\alpha^t:=1-\beta^t$, with $\beta^t$ represent the forward process variances.
$\epsilon^t$ denotes the added Gaussian noise, and $\mathbf{c}$ is the text condition, processed by the U-Net’s cross-attention module.

Moreover, the Denoising Diffusion Implicit Model \citep{song2020denoising} (DDIM) enables the inversion of the latent variable $z_0$ to $z_t$ in a deterministic manner. The formula is as follows:
\begin{equation}
\label{eq:ddim}
     z^{t} = \sqrt{\frac{\alpha^{t}}{\alpha^{t-1}}}z^{t-1}  +  + \left(\sqrt{\frac{1}{\alpha^{t}} - 1} - \sqrt{\frac{1}{\alpha^{t-1}} - 1} \right) \cdot \epsilon_\theta\left(z^{t-1}, t-1, c, c'\right),\\
\end{equation}
where $c'$ represents the additional image prompt. 
In our implementation, we utilize the latent space features at $t$-th time step and $l$-th layer of U-Net from DDIM inversion for landmark matching.

\subsection{FaceShot: Bring any Character into Life} 
\label{sec:method}
\begin{wrapfigure}{r}{0.5\linewidth}
		\centering
        \vspace{-4mm}
		\includegraphics[width=\linewidth]{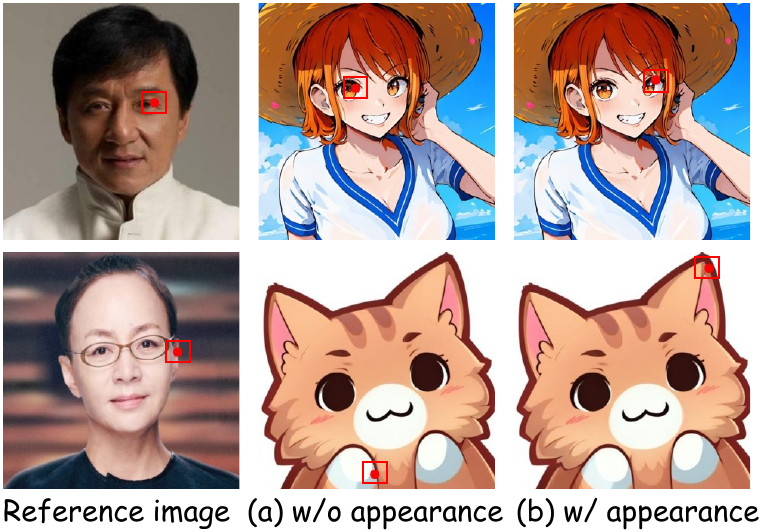}
\caption{Visualizations of point matching with (w/, highlighted in a red box) or without (w/o) appearance guidance using an anime diffusion model.}
\vspace{-4mm}
\label{fig:lora}
\end{wrapfigure}
Reviewing the remarkable variance in performance caused by inaccurate or accurate facial landmarks in Figure \ref{fig:intro}.
A well-generalized landmark detector is necessary for landmark-driven portrait animation model to bring any character into life.
Prior landmark detectors \citep{yang2023unipose, xu2022pose, zhou2023star} have either curated more diverse public datasets or introduced new loss functions during training to improve the generalization of landmark detection. 
However, within a supervised training paradigm, these detectors struggle to generalize to non-human characters, resulting inaccurate results for portrait animation. 
To address this, we propose an appearance-guided landmark matching module that generalizes to any character to generate precise landmarks.
In addition, to capture the subtle movements in driving videos, we offer a coordinate-based landmark retargeting module.
Finally, a character animation model is employed as our base model to animate reference characters.

\noindent{\textbf{Appearance-guided Landmark Matching.}} 
\cite{tang2023emergent, hedlin2024unsupervised, luo2024diffusion} demonstrate the strong semantic correspondence and generalization between diffusion features, where simple feature matching can map the point $p'$ on the reference image $I_{ref}$ to a semantic similar point $p$ on the target image $I_{tar}$. 
However, appearance discrepancies across different domains often result in mismatches, as shown in Figure \ref{fig:lora} (a), where the points on the left eye and right ear are incorrectly matched. 
A natural solution is to inject prior appearance knowledge through inference using a domain-specific diffusion model.
As shown in Figure \ref{fig:lora} (b), points are correctly matched when inferred on an anime diffusion model.

Since fine-tuning a diffusion model for each reference image is costly, inspired by IP-Adapter~\citep{ye2023ip}, we utilize image prompts to provide appearance guidance. 
Specifically, we treat the reference image $I_{ref}$ and the target image $I_{tar}$ as image prompts, denoted as $c'_{ref}$ and $c'_{tar}$.
We then apply the DDIM inversion process to obtain deterministic diffusion features $f_{ref}$ and $f_{tar}$ from $I_{ref}$ and $I_{tar}$ at time step $t$ and the $l$-th layer of the U-Net:
\begin{equation}
\begin{aligned}
    f_{tar} &= F_l(\epsilon_\theta(z^{t-1}_{tar}, t-1, c, c'_{ref})), \\
    f_{ref} &= F_l(\epsilon_\theta(z^{t-1}_{ref}, t-1, c, c'_{tar})), 
\end{aligned}
\end{equation}
where $z^{t-1}_{tar}$ and $z^{t-1}_{ref}$ are iteratively sampled by Eq. \ref{eq:ddim} from $z^{0}_{tar} = \mathcal{E}(I_{tar})$ and $z^{0}_{ref} = \mathcal{E}(I_{ref})$ using the text prompt $c=$\textit{``a photo of a face''} and the image prompts $c'_{ref}$ and $c'_{tar}$, respectively. $F_l$ denotes the function that extracts the output feature at the $l$-th layer of the U-Net.

After obtaining the diffusion features $f_{ref} \in \mathbb{R}^{1 \times C_l \times h_l \times w_l}$ and $f_{tar} \in \mathbb{R}^{1 \times C_l \times h_l \times w_l}$, we upsample them to $f_{ref}' \in \mathbb{R}^{1 \times C_l \times H \times W}$ and $f_{tar}' \in \mathbb{R}^{1 \times C_l \times H \times W}$ to match the resolutions of $I_{ref}$ and $I_{tar}$.
To improve performance and stability, we construct the average feature of the $i$-th landmark point $p_{tar}^i$ from $k$ target images to match the corresponding point $p^i_{ref}$ in the reference image, as follows:
\begin{equation}
\label{eq:opt}
    p^i_{ref} = \underset{p_{ref}}{\arg\min} \ d_{cos}\left(\frac{1}{k}\sum_{j=1}^k f'^j_{tar}(p_{tar}^{j,i}), f'_{ref}(p_{ref})\right),
\end{equation}

where $f'(p) \in \mathbb{R}^{1\times C_l}$ represents the diffusion feature vector at point $p$, $d_{cos}$ denotes the cosine distance and $p_{ref}$ refers to points in the reference feature map.
Finally, we denote the matched landmark points of the reference image as $L^0_{ref} = \{p^i_{ref} \mid i = 1, \dots, N\}$, where $N$ represents the number of facial landmarks.

\begin{wrapfigure}{r}{0.5\linewidth}
		\centering
        \vspace{-4mm}
		\includegraphics[width=\linewidth]{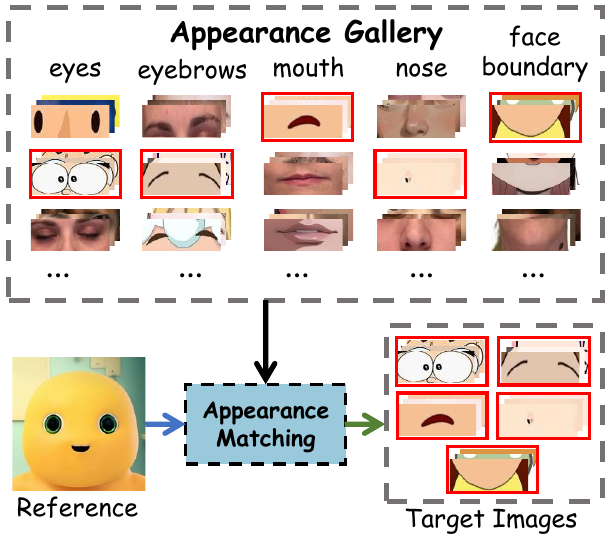}
\caption{Illustration of our appearance gallery. We output the closest domains for each reference image to reduce the appearance discrepancy.}
\vspace{-4mm}
\label{fig:gallery}
\end{wrapfigure}
\noindent{\textbf{Appearance Gallery.}} 
We introduce an appearance gallery $G = \left[G_e, G_m, G_n, G_{eb}, G_{fb}\right]$, which is a collection of five prior components—\textit{eyes, mouth, nose, eyebrows}, and \textit{face boundary}—across various domains, with each domain containing $k$ images. For a reference image $I_{ref}$, we reconstruct the target image as $I_{tar} = [G_e^*, G_m^*, G_n^*, G_{eb}^*, G_{fb}^*]$ by matching $I_{ref}$ with the closest domain in the appearance gallery $G$, thereby explicitly reducing the appearance discrepancy between the reference and target images, as shown in Figure \ref{fig:gallery}.

\begin{wrapfigure}{r}{0.6\linewidth}
		\centering
        \vspace{-4mm}
		\includegraphics[width=\linewidth]{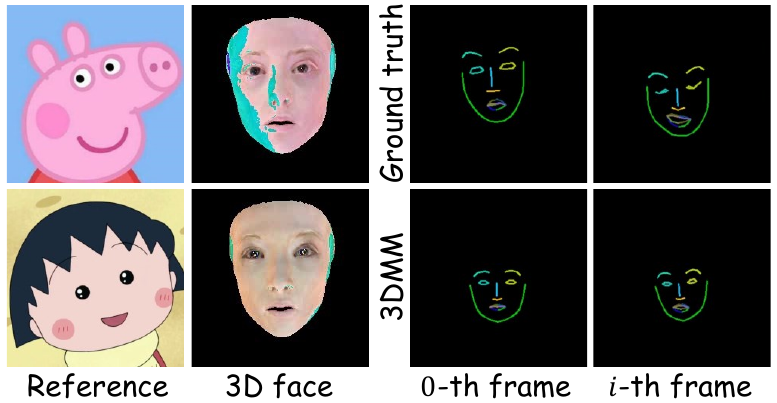}
\caption{Visualizations of 3D face and retargeting results using 3DMM.}
\vspace{-4mm}
\label{fig:3dmm}
\end{wrapfigure}

\noindent{\textbf{Coordinate-based Landmark Retargeting.}} 
Currently, \cite{niu2024mofa, wei2024aniportrait, ma2024follow} utilize 3D Morphable Models (3DMM)~\citep{booth20163d} to generate the landmark sequence of the reference image by applying 3D face parameters. However, 3DMM-based methods often struggle to generalize to non-human character faces due to the limited number of high-quality 3D data and their inability to capture subtle expression movements~\citep{retsinas20243d}. 
As shown in Figure \ref{fig:3dmm}, the head shapes of the 3D face are not well aligned with the input images, and subtle movements, such as eye closures, are absent in the $i$-th frame.
Therefore, we propose a coordinate-based landmark retargeting module designed to generate a retargeted landmark sequence $L_{ref}$ that can stably capture the subtle movements from driving video based on transformations in rectangular coordinate systems.

Our module consists of two stages, which respectively retarget the global motion of entire face and the local motion of different facial parts from driving sequence to reference image.
In the first stage, the global motion from the $0$-th to the $m$-th driving frame is defined as the translation $\Delta O_{dri}=O_{dri}^m - O_{dri}^0$ and rotation $\Delta \theta_{dri}=\theta_{dri}^m - \theta_{dri}^0$ of the corresponding global rectangular coordinate systems $(O^0_{dri}, \theta^0_{dri})$ and $(O^m_{dri}, \theta^m_{dri})$.
Specifically, the global rectangular coordinate system is constructed by the origin $O$ and angle $\theta$, which are calculated from the endpoints of face boundary.
Afterward, the global coordinate system of reference image at $m$-th frame can be calculated from those of $0$-th frame's as follows:
\begin{equation}
        O_{ref}^m = O_{ref}^0 + \Delta O_{dri}, \ \theta_{ref}^m = \theta_{ref}^0 + \Delta \theta_{dri}.
\end{equation}
Finally, we transfer the coordinates of the landmark points from $(O_{ref}^0, \theta_{ref}^0)$ to $(O_{ref}^m, \theta_{ref}^m)$, representing the global motion of the entire reference face.

\begin{figure*}[t]
    \centering
    \includegraphics[width=1.0\linewidth]{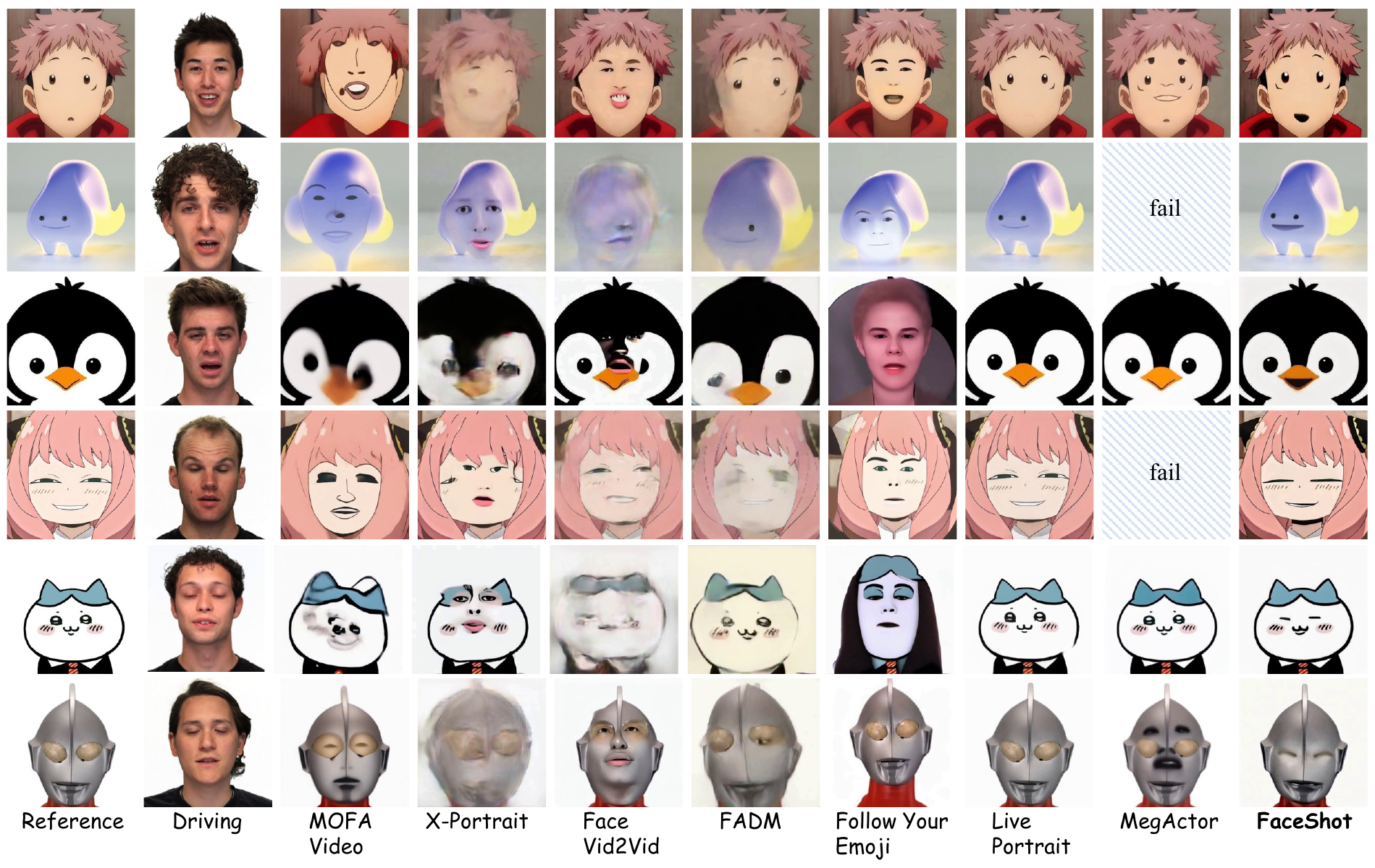}
    \caption{Qualitative comparison with SOTA portrait animation methods. Slash boxes represent that the method has fail to generate animation for this character.}
    \vspace{-4mm}
    \label{fig:visual}
\end{figure*}

In stage two, the local motion involves two processes: the relative motion and point motion, applied to five facial parts: eyes, mouth, nose, eyebrows, and face boundary.
The relative motion is similar to the global motion, but the part-specific coordinate systems are calculated from the endpoints of each part. 
Furthermore, to constrain each part within a reasonable facial range, we scale $\Delta O_{dri}$ as $\frac{b_{ref}}{b_{dri}}\Delta O_{dri}$, where $b$ represents the distance from the origin to the boundary of each part.
Next, we model the point motion as follows:
\begin{equation}
    p^{m,i}_{ref} = \left( \frac{p^{m,i}_{dri}[0]}{p^{0,i}_{dri}[0]} \cdot p^{0,i}_{ref}[0], \frac{p^{m,i}_{dri}[1]}{p^{0,i}_{dri}[1]} \cdot p^{0,i}_{ref}[1] \right),
\end{equation}
where $p^{m,i}_{ref}$ and $p^{m,i}_{dri}$ represent the coordinates in the part-specific coordinate system for the $m$-th frame and $i$-th point.
This simple yet effective design enables us to capture both global and local, obvious and subtle motions into landmark sequence $L_{ref} = \{L_{ref}^j \mid j=1, \dots, M\}$ for any character, where $M$ represents the number of video frames.

\noindent{\textbf{Character Animation Model.}} 
\label{sec:animation}
After obtaining the reference landmark sequence $L_{ref}$, it can be applied to any landmark-driven animation model to animate the character portrait. Specifically, $L_{ref}$ is treated as an additional condition for the U-Net, either injected via a ControlNet-like structure~\citep{niu2024mofa} or incorporated directly into the latent space~\citep{hu2024animate, wei2024aniportrait}. 
This enables the model to precisely track the motion encoded in the landmark sequence while preserving the character’s visual identity. 
Moreover, this flexible condition can be seamlessly extended to various architectures, enhancing scalability across diverse animation tasks.

\section{Experiments}

\subsection{Implement Detail}
\label{sec:exp}
In this work, we employ MOFA-Video~\citep{niu2024mofa}, a Stable Video Diffusion \citep{blattmann2023stable} (SVD)-based landmark-driven animation model, as our base character animation model.
For appearance-guided landmark matching, we utilize Stable Diffusion v1.5 along with the pre-trained weights of IP-Adapter~\citep{ye2023ip} to extract diffusion features from the images. Specifically, we set the time step $t=301$, the U-Net layer $l=6$, and the number of target images $k=10$.
Additionally, following MOFA-Video, we use $N=68$ keypoints \citep{sagonas2016300} as facial landmarks and $M=64$ frames for animation.
More details are shown in Appendix.

\begin{wrapfigure}{r}{0.5\linewidth}
		\centering
  \vspace{-4mm}
		\includegraphics[width=\linewidth]{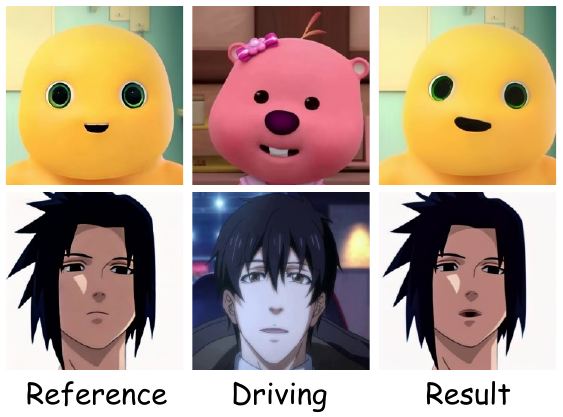}
\caption{Visualizations of character animation from non-human driving videos.}
\vspace{-4mm}
\label{fig:ood}
\end{wrapfigure}
\noindent{\textbf{Evaluation Metrics.}} 
Following \cite{xie2024x, ma2024follow}, we employ four metrics to evaluate identity similarity, high- and low-level image quality and expression accuracy.
Specifically, we utilize ArcFace score \citep{deng2019arcface} that calculates average cosine similarity between source and generated videos as identity similarity.
We also employ HyperIQA \citep{zhang2023blind} and LAION Aesthetic \citep{schuhmann2022laion} for evaluating image quality from low- and high-level.
Moreover, we conduct the expression evaluation following the steps of Point-Tracking in MimicBench\footnote{https://github.com/open-mmlab/MimicBench}.

\noindent{\textbf{Character Benchmark.}} 
To comprehensively evaluate the effectiveness and generalization ability of portrait animation methods towards characters, we build CharacBench that comprises 46 characters from various domains, such as animals, emojis, toys and anime characters.
Characters in CharacBench are collected from Internet by following the guideline of ensuring that the characters do not resemble human facial features as much as possible. 
Moreover, we consider videos of human head from RAVDESS \citep{livingstone2018ryerson} as our driving videos.

\subsection{Comparison with SOTA Methods}
\label{sec:comparison}

\noindent{\textbf{Qualitative  Results.}} 
We compare proposed {FaceShot} with SOTA portrait animation methods, including {MOFA-Video} \citep{niu2024mofa}, {X-Portrait} \citep{xie2024x}, {FaceVid2Vid} \citep{wang2021one}, {FADM} \citep{zeng2023face}, {Follow Your Emoji} \citep{ma2024follow}, {LivePortrait} \citep{guo2024liveportrait}, and {MegActor} \citep{yang2024megactor}. 
Visual comparisons are presented in Figure \ref{fig:visual}, where \textit{fail} indicates that the method was unable to generate animation for the character.
As {AniPortrait} \citep{wei2024aniportrait} fails with most non-human characters, we only provide its quantitative results. 
We observe that most methods such as MOFA-Video, X-Portrait, FaceVid2Vid and Follow Your Emoji, are influenced by the human prior in the driving video, resulting in human facial features appearing on character faces. 
In contrast, FaceShot effectively preserves the identity of reference characters through precise landmarks provided by our proposed appearance-guided landmark matching module.
Furthermore, while most methods struggle to retarget the motions like eye closure and mouth opening, our coordinated-based landmark retargeting module enables FaceShot to capture subtle movements.

Beyond its effective character animation capability, {FaceShot} can also animate reference characters from non-human driving videos, extending the application of portrait animation from human-related videos to any video as shown in Fig. \ref{fig:ood}.
This demonstrates its potential for open-domain portrait animations.

\begin{table*}[t]
\vspace{-4mm}
\caption{Quantitative comparison between FaceShot and other SOTA methods on CharacBench. The best result is marked in \textbf{bold}, and the second-best performance is highlighted in \underline{underline}. Symbol * indicates that there are some failure cases in these methods, we report the values of these methods \textit{only for reference}.
}
\centering
\footnotesize
\setlength\tabcolsep{5pt} 
\resizebox{1.0\linewidth}{!}{
\begin{tabular}{c|cccc|ccc}
\toprule
\multirow{2}{*}{Methods} & \multicolumn{4}{c|}{Metrics}  &  \multicolumn{3}{c}{User Preference}   \\ \cline{2-8}
& ArcFace $\uparrow$ & HyperIQA $\uparrow$ & Aesthetic $\uparrow$ & Point-Tracking $\downarrow$  & Motion $\uparrow$& Identity $\uparrow$& Overall $\uparrow$  \\
\midrule
FaceVid2Vid & 0.525 &33.721&4.267 & \underline{6.944} & 3.58  &3.83 & 4.52  \\
FADM & {0.633} &39.402&4.522 & 6.993 &1.93  &2.04 & 1.96 \\
X-Portrait  & 0.490 & \underline{52.357} &4.754&7.301& 1.47 & 1.63 & 1.57   \\
Follow Your Emoji & 0.612 & 52.056&{4.906}& {6.960} & 6.91 &6.67 & 6.74  \\
\hline
AniPortrait* & 0.634 &{55.951}&4.928&  {6.367} &5.84 & 5.64 & 5.39  \\
MegActor*  & 0.613 & 40.191 &4.855&7.183 & 6.53 & 6.75 &  6.26   \\
LivePortrait*  & {0.893} & {53.587} & {5.092}& 7.474 &\underline{7.33} &\underline{7.08} & \underline{7.11}    \\
\hline
MOFA-Video & \underline{0.695} & 52.272 &\underline{4.952}& 14.985 &3.27  & 3.04 & 3.18   \\
\textbf{FaceShot}   & \textbf{0.848} & \textbf{53.723} & \textbf{5.036} & \textbf{6.935}&\textbf{8.14}  &\textbf{8.32}  &\textbf{8.27} \\ 
\bottomrule
\end{tabular}
}
\vspace{-4mm}
\label{tab:quantative}
\end{table*}

\begin{wrapfigure}{r}{0.6\linewidth}
\centering
\vspace{-4mm}
    \includegraphics[width=\linewidth]{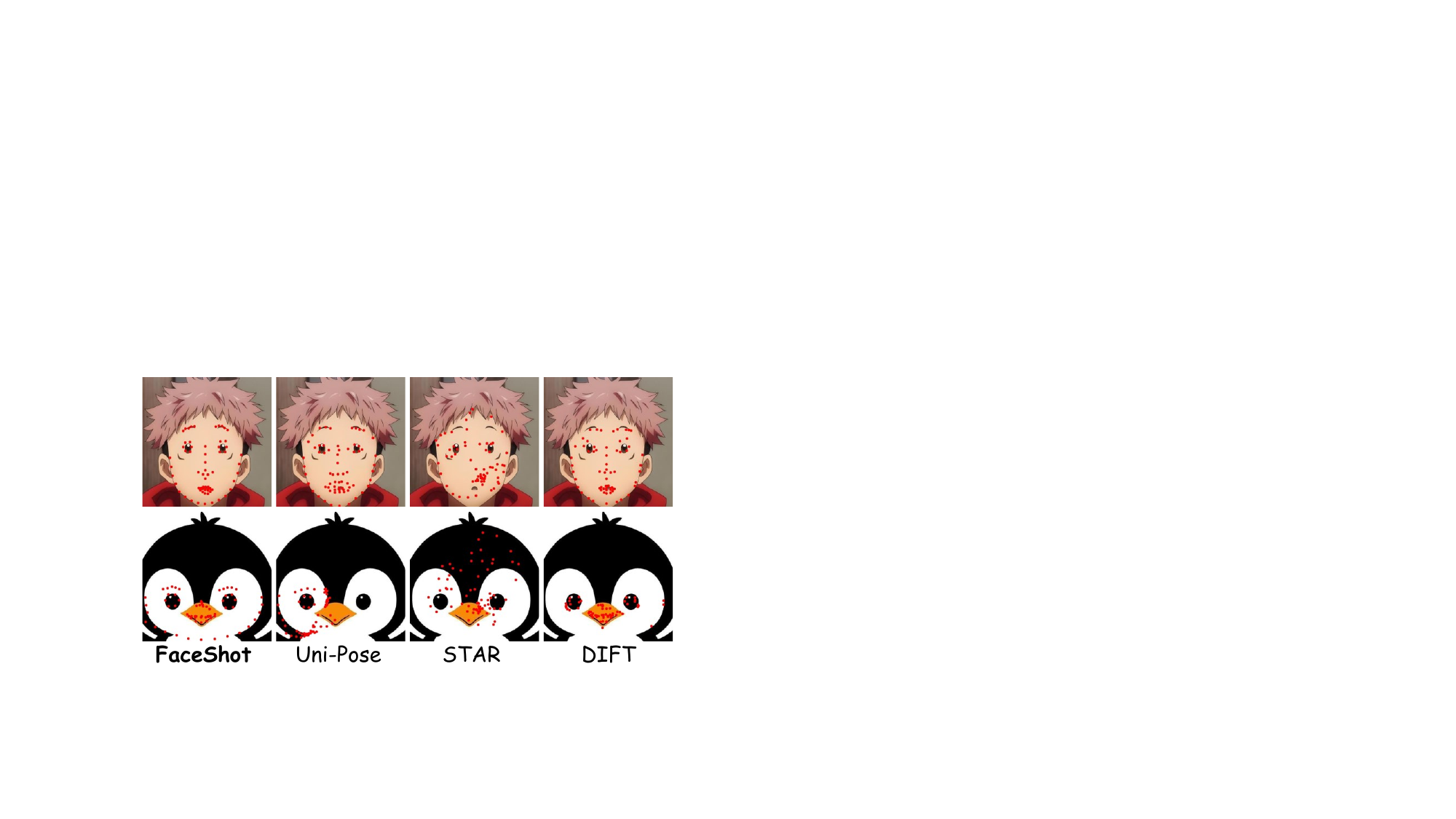}
    \caption{The visualizations of landmarks detection on different characters in CharacBench using DIFT, STAR, Uni-Pose and FaceShot.}
    \vspace{-4mm}
    \label{fig:character}
\end{wrapfigure}
\noindent{\textbf{Quantitative Results.}} 
We conduct a quantitative comparison on the metrics mentioned in Section \ref{sec:exp}. 
Please note that some methods, such as LivePortrait, MegActor, and AniPortrait, fail to generate animations for certain characters when they are unable to detect the face.
Therefore, for a fair comparison, we report the \textbf{failure rate} for these methods as follows: AniPortrait (39.13\%), MegActor (36.50\%), and LivePortrait (16.67\%), and we calculate their metric values on successful characters \textbf{only for reference purposes}. 
Based on Table \ref{tab:quantative}, FaceShot demonstrates significantly superior performance across various metrics compared to other methods on CharacBench.
Specifically, FaceShot achieves the highest score in terms of {ArcFace} (0.848), demonstrating the effectiveness of the precise landmarks generated by the appearance-guided landmark matching module in preserving facial identity. 
{FaceShot} achieves superior {HyperIQA} (53.723) and {Aesthetic} (5.036) scores, indicating better image quality.
Additionally, the coordinate-based landmark retargeting module contributes to the competitive {point tracking} score (6.935), highlighting its ability to handle motion effectively. 
It is important to note that our method has achieved significant improvements across all metrics compared to the base method, {MOFA-Video}, further demonstrating the effectiveness of our proposed {FaceShot}.

\noindent{\textbf{User Preference.}} 
Additionally, we randomly selected 15 case examples and enlisted 20 volunteers to evaluate each method across three key dimensions: Motion, Identity, and Overall User Satisfaction. Volunteers ranked the animations based on these criteria, ensuring a fair and comprehensive comparison between the methods.
As shown in Table \ref{tab:quantative}, FaceShot achieves the highest scores in Motion, Identity, and Overall categories, demonstrating its robust animation capabilities across diverse characters and driving videos.

\subsection{Ablation Studies}
\label{sec:abla}

\noindent{\textbf{Appearance-guided Landmark Matching.}} 
To evaluate the effectiveness of our appearance-guided landmark matching module, we compare it with SOTA unsupervised algorithm DIFT \citep{tang2023emergent} and supervised algorithms Uni-Pose \citep{yang2023unipose} and STAR \citep{zhou2023star} on CharacBench.
Specifically, we recruited volunteers to annotate the landmarks of images in CharacBench as ground-truth values for calculating the corresponding Normalized Mean Error (NME) value. 

\begin{table}[h]
    \centering
    \caption{Albation studies of our appearance-guided landmark matching module with supervised SOTA methods Uni-Pose and STAR and unsupervised method DIFT on CharacBench. Best result is marked in \textbf{bold}, and the second-best performance is highlighted in \underline{underline}.}
    \begin{tabular}{c|cccc}
    \toprule
        Methods &  NME $\downarrow$ & ArcFace $\uparrow$  & HyperIQA $\uparrow$ & Aesthetic $\uparrow$ \\
         \midrule
        STAR  & 24.530 & 52.849 & 0.829 & 4.989\\
        Uni-Pose  & 13.731 & \underline{53.685} & \textbf{0.851} & \underline{5.025} \\
        DIFT  & \underline{11.448} & 53.506 &  0.843 & 5.023\\
        FaceShot & \textbf{8.569} & \textbf{53.723} & \underline{0.848} & \textbf{5.036}\\
         \bottomrule
    \end{tabular}
    \label{tab:landmarks}
\end{table}

As illustrated in Table \ref{tab:landmarks}, FaceShot achieves the lowest NME on CharacBench. 
The visual landmark results are shown in Figure \ref{fig:character}, FaceShot accurately detects the landmarks on the non-human characters, whereas others fail to match the positions of the eyes and mouth.
Furthermore, we observe that Faceshot also achieves the best scores in Table \ref{tab:landmarks}, highlighting the necessity of precise landmarks of landmark-driven portrait animation models.

\begin{wrapfigure}{r}{0.7\linewidth}
    \centering
        \vspace{-4mm}
    \includegraphics[width=\linewidth]{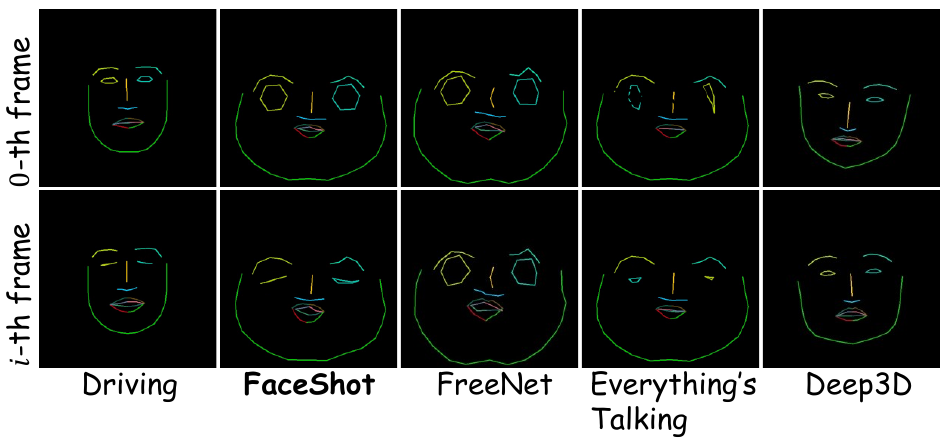}
    \caption{Visual results of landmark retargeting between FaceShot and other methods.}
    \vspace{-4mm}
    \label{fig:transfer}
\end{wrapfigure}
\noindent{\textbf{Coordinate-based Landmark Retargeting.}} 
To evaluate the effectiveness of our landmark retargeting module, we provide a comparison with SOTA landmark retargeting methods, including Deep3D (3DMM) \citep{deng2019accurate}, Everthing's Talking \citep{song2021everything} and FreeNet \citep{zhang2020freenet}, where Everthing's Talking models the B\'{e}zier Curve as the motion controller and FreeNet trains a parameterized network for retargeting.

\begin{table}[h]
    \centering
    \caption{Albation studies of coordinate-based landmark retargeting module with SOTA methods Deep3D, Everthing's Talking and FreeNet. Best result is marked in \textbf{bold}.}
    \setlength\tabcolsep{3.5pt} 
    \begin{tabular}{c|cccc}
    \toprule
        Metric & FaceShot & Deep3D & \makecell{Everthing's \\ Talking} & FreeNet \\
         \midrule
        \makecell{Point-\\ Tracking} $\downarrow$  & \textbf{6.935} & 8.282 & 8.382 & 8.272\\
         \bottomrule
    \end{tabular}
    \label{tab:motion}
\end{table}
As shown in Figure \ref{fig:transfer}, due to the precision loss of fitting a Bezier curve, Everthing's Talking always generates the inaccurate retargeting.
FreeNet and 3DMM have strict requirements for the distribution facial features, making it unable to adapt to non-human characters.
In contrast, our module can precisely capture the subtle motion such as mouth opening, eye closure, and global face movement.
Additionally, FaceShot also achieves the lowest point-tracking score in Table \ref{tab:motion}, demonstrating its effectiveness in capturing the consistent face movements.

\begin{wrapfigure}{r}{0.6\linewidth}
    \centering
    \vspace{-4mm}
    \includegraphics[width=\linewidth]{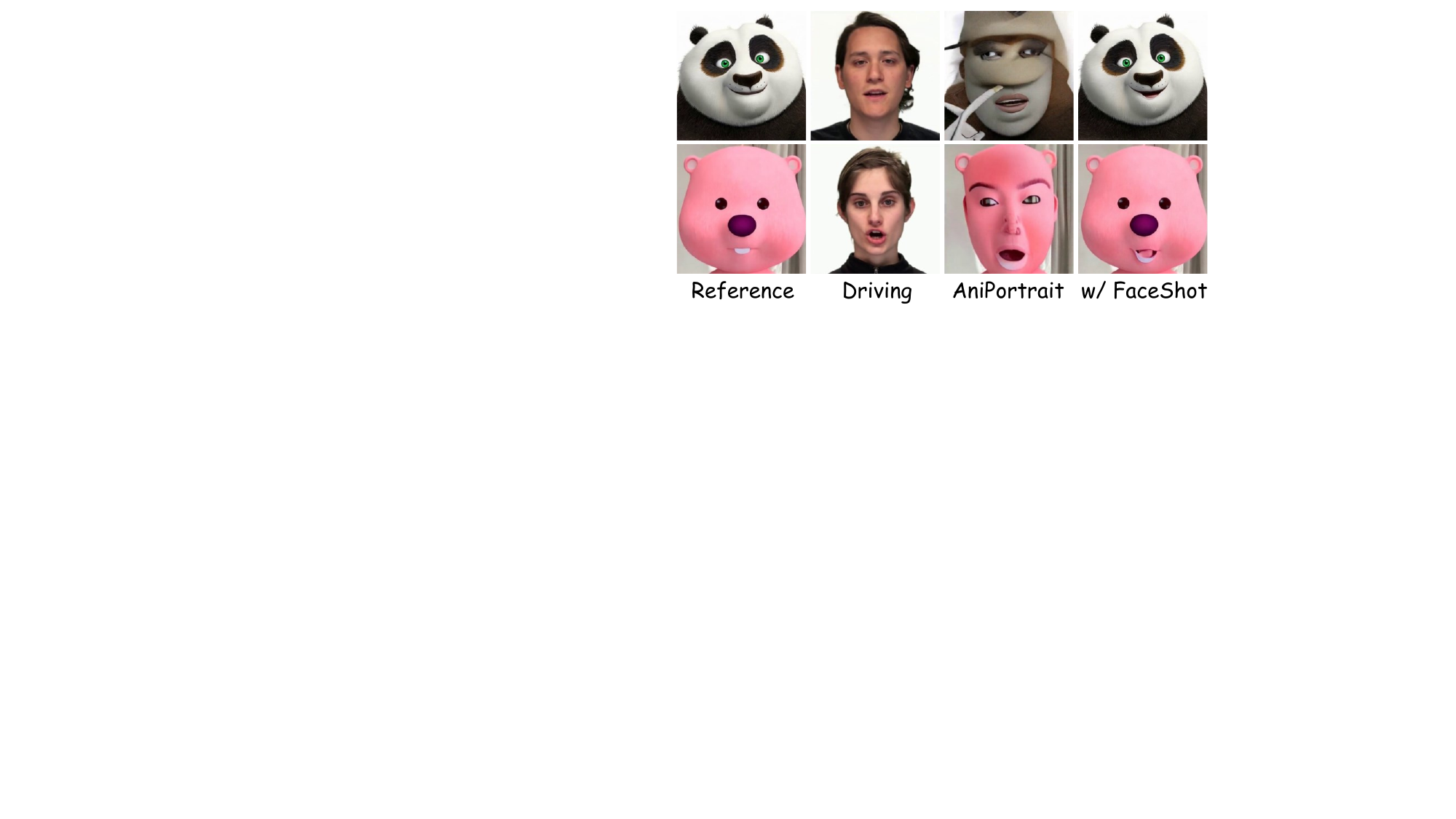}
    \caption{Visual results of AniPortrait with or without FaceShot as a plugin to generate animation results.}
    \vspace{-4mm}
    \label{fig:anip}
\end{wrapfigure}
\noindent{\textbf{As a Plugin.}} 
Experiments in Section \ref{sec:comparison} have demonstrated that FaceShot can significantly improve the performance of the landmark-driven method MOFA-Video~\citep{niu2024mofa}.
To further verify its effectiveness as a plugin for landmark-driven animation models, we input the reposed landmark sequences generated by FaceShot into AniPortrait's~\citep{wei2024aniportrait} pipeline.
As shown in Figure \ref{fig:anip}, inaccurate landmarks in original AniPortrait often result in distortions, producing results that resemble real humans.
In contrast, FaceShot can provide precise and robust landmarks for characters, leading to harmonious and stable animation results.

\section{Conclusion}
In this paper, we introduced FaceShot, a training-free portrait animation framework that animates any character from any driven video.
By leveraging semantic correspondence in latent diffusion model features, FaceShot addresses the limitations of existing landmark-driven methods, enabling precise landmark matching and landmark retargeting.
This powerful capability not only extends the application of portrait animation beyond traditional boundaries but also enhances the realism and consistency of animations in landmark-driven models. 
FaceShot is also compatible with any landmark-driven animation model as a plugin.
Additionally, experimental results on CharacBench, a benchmark featuring diverse characters, demonstrate that FaceShot consistently outperforms current SOTA methods.

\noindent{\textbf{Future Work.}} 
Although FaceShot shows strong performance, future work could focus on enhancing appearance-guided landmark matching by refining semantic feature extraction from latent diffusion models, particularly for complex facial geometries. Furthermore, parameterizing landmark retargeting could offer more precise control over facial expressions, improving the adaptability of FaceShot across diverse character types and styles.

\section*{Ethics Statement}
In developing FaceShot, a training-free portrait animation framework that animates any characters from any driven video, we are dedicated to upholding ethical standards and promoting responsible AI use. We acknowledge potential risks, such as deepfake misuse or unauthorized media manipulation, and stress the importance of applying this technology in ways that respect privacy, consent, and individual rights.
Our code will be publicly released to encourage responsible use in areas like entertainment and education, while discouraging unethical practices, including misinformation and harassment. We also advocate for continued research on safeguards and detection mechanisms to prevent misuse and ensure adherence to ethical guidelines and legal frameworks.

\section*{Acknowledgments}
This project is supported by the National Key R\&D Program of China (No. 2022ZD0161600) and the National Natural Science Fund of China (No. 62473286).

\newpage
\bibliography{iclr2025_conference}

\begin{thebibliography}{78}
\providecommand{\natexlab}[1]{#1}
\providecommand{\url}[1]{\texttt{#1}}
\expandafter\ifx\csname urlstyle\endcsname\relax
  \providecommand{\doi}[1]{doi: #1}\else
  \providecommand{\doi}{doi: \begingroup \urlstyle{rm}\Url}\fi

\bibitem[Blattmann et~al.(2023{\natexlab{a}})Blattmann, Dockhorn, Kulal, Mendelevitch, Kilian, Lorenz, Levi, English, Voleti, Letts, et~al.]{blattmann2023stable}
Andreas Blattmann, Tim Dockhorn, Sumith Kulal, Daniel Mendelevitch, Maciej Kilian, Dominik Lorenz, Yam Levi, Zion English, Vikram Voleti, Adam Letts, et~al.
\newblock Stable video diffusion: Scaling latent video diffusion models to large datasets.
\newblock \emph{arXiv preprint arXiv:2311.15127}, 2023{\natexlab{a}}.

\bibitem[Blattmann et~al.(2023{\natexlab{b}})Blattmann, Rombach, Ling, Dockhorn, Kim, Fidler, and Kreis]{blattmann2023align}
Andreas Blattmann, Robin Rombach, Huan Ling, Tim Dockhorn, Seung~Wook Kim, Sanja Fidler, and Karsten Kreis.
\newblock Align your latents: High-resolution video synthesis with latent diffusion models.
\newblock In \emph{Proceedings of the IEEE/CVF Conference on Computer Vision and Pattern Recognition}, pp.\  22563--22575, 2023{\natexlab{b}}.

\bibitem[Booth et~al.(2016)Booth, Roussos, Zafeiriou, Ponniah, and Dunaway]{booth20163d}
James Booth, Anastasios Roussos, Stefanos Zafeiriou, Allan Ponniah, and David Dunaway.
\newblock A 3d morphable model learnt from 10,000 faces.
\newblock In \emph{Proceedings of the IEEE conference on computer vision and pattern recognition}, pp.\  5543--5552, 2016.

\bibitem[Cootes et~al.(2000)Cootes, Baldock, and Graham]{cootes2000introduction}
Tim Cootes, ER~Baldock, and J~Graham.
\newblock An introduction to active shape models.
\newblock \emph{Image processing and analysis}, 328:\penalty0 223--248, 2000.

\bibitem[Cootes et~al.(2001)Cootes, Edwards, and Taylor]{cootes2001active}
Timothy~F. Cootes, Gareth~J. Edwards, and Christopher~J. Taylor.
\newblock Active appearance models.
\newblock \emph{IEEE Transactions on pattern analysis and machine intelligence}, 23\penalty0 (6):\penalty0 681--685, 2001.

\bibitem[Dai et~al.(2023)Dai, Zhang, Yao, Qiu, Zhu, Qin, and Wang]{dai2023animateanything}
Zuozhuo Dai, Zhenghao Zhang, Yao Yao, Bingxue Qiu, Siyu Zhu, Long Qin, and Weizhi Wang.
\newblock Animateanything: Fine-grained open domain image animation with motion guidance.
\newblock \emph{arXiv e-prints}, pp.\  arXiv--2311, 2023.

\bibitem[Deng et~al.(2019{\natexlab{a}})Deng, Guo, Xue, and Zafeiriou]{deng2019arcface}
Jiankang Deng, Jia Guo, Niannan Xue, and Stefanos Zafeiriou.
\newblock Arcface: Additive angular margin loss for deep face recognition.
\newblock In \emph{Proceedings of the IEEE/CVF conference on computer vision and pattern recognition}, pp.\  4690--4699, 2019{\natexlab{a}}.

\bibitem[Deng et~al.(2019{\natexlab{b}})Deng, Yang, Xu, Chen, Jia, and Tong]{deng2019accurate}
Yu~Deng, Jiaolong Yang, Sicheng Xu, Dong Chen, Yunde Jia, and Xin Tong.
\newblock Accurate 3d face reconstruction with weakly-supervised learning: From single image to image set.
\newblock In \emph{Proceedings of the IEEE/CVF conference on computer vision and pattern recognition workshops}, pp.\  0--0, 2019{\natexlab{b}}.

\bibitem[Doll{\'a}r et~al.(2010)Doll{\'a}r, Welinder, and Perona]{dollar2010cascaded}
Piotr Doll{\'a}r, Peter Welinder, and Pietro Perona.
\newblock Cascaded pose regression.
\newblock In \emph{2010 IEEE Computer Society Conference on Computer Vision and Pattern Recognition}, pp.\  1078--1085. IEEE, 2010.

\bibitem[Donahue et~al.(2016)Donahue, Kr{\"a}henb{\"u}hl, and Darrell]{donahue2016adversarial}
Jeff Donahue, Philipp Kr{\"a}henb{\"u}hl, and Trevor Darrell.
\newblock Adversarial feature learning.
\newblock \emph{arXiv preprint arXiv:1605.09782}, 2016.

\bibitem[Drobyshev et~al.(2022)Drobyshev, Chelishev, Khakhulin, Ivakhnenko, Lempitsky, and Zakharov]{drobyshev2022megaportraits}
Nikita Drobyshev, Jenya Chelishev, Taras Khakhulin, Aleksei Ivakhnenko, Victor Lempitsky, and Egor Zakharov.
\newblock Megaportraits: One-shot megapixel neural head avatars.
\newblock In \emph{Proceedings of the 30th ACM International Conference on Multimedia}, pp.\  2663--2671, 2022.

\bibitem[Feng et~al.(2021)Feng, Feng, Black, and Bolkart]{feng2021learning}
Yao Feng, Haiwen Feng, Michael~J Black, and Timo Bolkart.
\newblock Learning an animatable detailed 3d face model from in-the-wild images.
\newblock \emph{ACM Transactions on Graphics (ToG)}, 40\penalty0 (4):\penalty0 1--13, 2021.

\bibitem[Gao et~al.(2024)Gao, Liu, Sun, Tang, Zeng, Chen, and Zhao]{gao2024styleshot}
Junyao Gao, Yanchen Liu, Yanan Sun, Yinhao Tang, Yanhong Zeng, Kai Chen, and Cairong Zhao.
\newblock Styleshot: A snapshot on any style.
\newblock \emph{arXiv preprint arXiv:2407.01414}, 2024.

\bibitem[Gong et~al.(2024)Gong, Zhu, Li, Kang, Wang, Ge, and Zheng]{gong2024atomovideo}
Litong Gong, Yiran Zhu, Weijie Li, Xiaoyang Kang, Biao Wang, Tiezheng Ge, and Bo~Zheng.
\newblock Atomovideo: High fidelity image-to-video generation.
\newblock \emph{arXiv preprint arXiv:2403.01800}, 2024.

\bibitem[Goodfellow et~al.(2020)Goodfellow, Pouget-Abadie, Mirza, Xu, Warde-Farley, Ozair, Courville, and Bengio]{gan}
Ian Goodfellow, Jean Pouget-Abadie, Mehdi Mirza, Bing Xu, David Warde-Farley, Sherjil Ozair, Aaron Courville, and Yoshua Bengio.
\newblock Generative adversarial networks.
\newblock \emph{Communications of the ACM}, 63\penalty0 (11):\penalty0 139--144, 2020.

\bibitem[Guo et~al.(2020)Guo, Zhu, Yang, Yang, Lei, and Li]{guo2020towards}
Jianzhu Guo, Xiangyu Zhu, Yang Yang, Fan Yang, Zhen Lei, and Stan~Z Li.
\newblock Towards fast, accurate and stable 3d dense face alignment.
\newblock In \emph{European Conference on Computer Vision}, pp.\  152--168. Springer, 2020.

\bibitem[Guo et~al.(2024)Guo, Zhang, Liu, Zhong, Zhang, Wan, and Zhang]{guo2024liveportrait}
Jianzhu Guo, Dingyun Zhang, Xiaoqiang Liu, Zhizhou Zhong, Yuan Zhang, Pengfei Wan, and Di~Zhang.
\newblock Liveportrait: Efficient portrait animation with stitching and retargeting control.
\newblock \emph{arXiv preprint arXiv:2407.03168}, 2024.

\bibitem[Guo et~al.(2023)Guo, Yang, Rao, Liang, Wang, Qiao, Agrawala, Lin, and Dai]{guo2023animatediff}
Yuwei Guo, Ceyuan Yang, Anyi Rao, Zhengyang Liang, Yaohui Wang, Yu~Qiao, Maneesh Agrawala, Dahua Lin, and Bo~Dai.
\newblock Animatediff: Animate your personalized text-to-image diffusion models without specific tuning.
\newblock \emph{arXiv preprint arXiv:2307.04725}, 2023.

\bibitem[Hedlin et~al.(2024)Hedlin, Sharma, Mahajan, He, Isack, Kar, Rhodin, Tagliasacchi, and Yi]{hedlin2024unsupervised}
Eric Hedlin, Gopal Sharma, Shweta Mahajan, Xingzhe He, Hossam Isack, Abhishek Kar, Helge Rhodin, Andrea Tagliasacchi, and Kwang~Moo Yi.
\newblock Unsupervised keypoints from pretrained diffusion models.
\newblock In \emph{Proceedings of the IEEE/CVF Conference on Computer Vision and Pattern Recognition}, pp.\  22820--22830, 2024.

\bibitem[Ho et~al.(2020)Ho, Jain, and Abbeel]{ho2020denoising}
Jonathan Ho, Ajay Jain, and Pieter Abbeel.
\newblock Denoising diffusion probabilistic models.
\newblock In \emph{NeurIPS}, 2020.

\bibitem[Hong et~al.(2022)Hong, Zhang, Shen, and Xu]{hong2022depth}
Fa-Ting Hong, Longhao Zhang, Li~Shen, and Dan Xu.
\newblock Depth-aware generative adversarial network for talking head video generation.
\newblock In \emph{Proceedings of the IEEE/CVF conference on computer vision and pattern recognition}, pp.\  3397--3406, 2022.

\bibitem[Hu(2024)]{hu2024animate}
Li~Hu.
\newblock Animate anyone: Consistent and controllable image-to-video synthesis for character animation.
\newblock In \emph{Proceedings of the IEEE/CVF Conference on Computer Vision and Pattern Recognition}, pp.\  8153--8163, 2024.

\bibitem[Huang et~al.(2020)Huang, Deng, Shen, Zhang, and Ye]{huang2020propagationnet}
Xiehe Huang, Weihong Deng, Haifeng Shen, Xiubao Zhang, and Jieping Ye.
\newblock Propagationnet: Propagate points to curve to learn structure information.
\newblock In \emph{Proceedings of the IEEE/CVF Conference on Computer Vision and Pattern Recognition}, pp.\  7265--7274, 2020.

\bibitem[Kingma(2013)]{vae}
Diederik~P Kingma.
\newblock Auto-encoding variational bayes.
\newblock \emph{arXiv preprint arXiv:1312.6114}, 2013.

\bibitem[Kowalski et~al.(2017)Kowalski, Naruniec, and Trzcinski]{kowalski2017deep}
Marek Kowalski, Jacek Naruniec, and Tomasz Trzcinski.
\newblock Deep alignment network: A convolutional neural network for robust face alignment.
\newblock In \emph{Proceedings of the IEEE conference on computer vision and pattern recognition workshops}, pp.\  88--97, 2017.

\bibitem[Kumar et~al.(2020)Kumar, Marks, Mou, Wang, Jones, Cherian, Koike-Akino, Liu, and Feng]{kumar2020luvli}
Abhinav Kumar, Tim~K Marks, Wenxuan Mou, Ye~Wang, Michael Jones, Anoop Cherian, Toshiaki Koike-Akino, Xiaoming Liu, and Chen Feng.
\newblock Luvli face alignment: Estimating landmarks' location, uncertainty, and visibility likelihood.
\newblock In \emph{Proceedings of the IEEE/CVF Conference on Computer Vision and Pattern Recognition}, pp.\  8236--8246, 2020.

\bibitem[Li et~al.(2024{\natexlab{a}})Li, Zhao, Wang, and Lin]{li2024towards}
Jiaxing Li, Hongbo Zhao, Yijun Wang, and Jianxin Lin.
\newblock Towards photorealistic video colorization via gated color-guided image diffusion models.
\newblock In \emph{Proceedings of the 32nd ACM International Conference on Multimedia}, pp.\  10891--10900, 2024{\natexlab{a}}.

\bibitem[Li et~al.(2024{\natexlab{b}})Li, Tan, and Gou]{li2024cascaded}
Yaokun Li, Guang Tan, and Chao Gou.
\newblock Cascaded iterative transformer for jointly predicting facial landmark, occlusion probability and head pose.
\newblock \emph{International Journal of Computer Vision}, 132\penalty0 (4):\penalty0 1242--1257, 2024{\natexlab{b}}.

\bibitem[Livingstone \& Russo(2018)Livingstone and Russo]{livingstone2018ryerson}
Steven~R Livingstone and Frank~A Russo.
\newblock The ryerson audio-visual database of emotional speech and song (ravdess): A dynamic, multimodal set of facial and vocal expressions in north american english.
\newblock \emph{PloS one}, 13\penalty0 (5):\penalty0 e0196391, 2018.

\bibitem[Luo et~al.(2024)Luo, Dunlap, Park, Holynski, and Darrell]{luo2024diffusion}
Grace Luo, Lisa Dunlap, Dong~Huk Park, Aleksander Holynski, and Trevor Darrell.
\newblock Diffusion hyperfeatures: Searching through time and space for semantic correspondence.
\newblock \emph{Advances in Neural Information Processing Systems}, 36, 2024.

\bibitem[Ma et~al.(2024)Ma, Liu, Wang, Pan, He, Yuan, Zeng, Cai, Shum, Liu, et~al.]{ma2024follow}
Yue Ma, Hongyu Liu, Hongfa Wang, Heng Pan, Yingqing He, Junkun Yuan, Ailing Zeng, Chengfei Cai, Heung-Yeung Shum, Wei Liu, et~al.
\newblock Follow-your-emoji: Fine-controllable and expressive freestyle portrait animation.
\newblock \emph{arXiv preprint arXiv:2406.01900}, 2024.

\bibitem[Merget et~al.(2018)Merget, Rock, and Rigoll]{merget2018robust}
Daniel Merget, Matthias Rock, and Gerhard Rigoll.
\newblock Robust facial landmark detection via a fully-convolutional local-global context network.
\newblock In \emph{Proceedings of the IEEE conference on computer vision and pattern recognition}, pp.\  781--790, 2018.

\bibitem[Ni et~al.(2023)Ni, Shi, Li, Huang, and Min]{ni2023conditional}
Haomiao Ni, Changhao Shi, Kai Li, Sharon~X Huang, and Martin~Renqiang Min.
\newblock Conditional image-to-video generation with latent flow diffusion models.
\newblock In \emph{Proceedings of the IEEE/CVF conference on computer vision and pattern recognition}, pp.\  18444--18455, 2023.

\bibitem[Nichol et~al.(2021)Nichol, Dhariwal, Ramesh, Shyam, Mishkin, McGrew, Sutskever, and Chen]{nichol2021glide}
Alex Nichol, Prafulla Dhariwal, Aditya Ramesh, Pranav Shyam, Pamela Mishkin, Bob McGrew, Ilya Sutskever, and Mark Chen.
\newblock Glide: Towards photorealistic image generation and editing with text-guided diffusion models.
\newblock \emph{arXiv preprint arXiv:2112.10741}, 2021.

\bibitem[Niu et~al.(2024)Niu, Cun, Wang, Zhang, Shan, and Zheng]{niu2024mofa}
Muyao Niu, Xiaodong Cun, Xintao Wang, Yong Zhang, Ying Shan, and Yinqiang Zheng.
\newblock Mofa-video: Controllable image animation via generative motion field adaptions in frozen image-to-video diffusion model.
\newblock \emph{arXiv preprint arXiv:2405.20222}, 2024.

\bibitem[Odena et~al.(2017)Odena, Olah, and Shlens]{odena2017conditional}
Augustus Odena, Christopher Olah, and Jonathon Shlens.
\newblock Conditional image synthesis with auxiliary classifier gans.
\newblock In \emph{International conference on machine learning}, pp.\  2642--2651. PMLR, 2017.

\bibitem[Radford(2015)]{radford2015unsupervised}
Alec Radford.
\newblock Unsupervised representation learning with deep convolutional generative adversarial networks.
\newblock \emph{arXiv preprint arXiv:1511.06434}, 2015.

\bibitem[Radford et~al.(2021)Radford, Kim, Hallacy, Ramesh, Goh, Agarwal, Sastry, Askell, Mishkin, Clark, et~al.]{clip}
Alec Radford, Jong~Wook Kim, Chris Hallacy, Aditya Ramesh, Gabriel Goh, Sandhini Agarwal, Girish Sastry, Amanda Askell, Pamela Mishkin, Jack Clark, et~al.
\newblock Learning transferable visual models from natural language supervision.
\newblock In \emph{International conference on machine learning}, pp.\  8748--8763. PMLR, 2021.

\bibitem[Ramesh et~al.(2022)Ramesh, Dhariwal, Nichol, Chu, and Chen]{ramesh2022hierarchical}
Aditya Ramesh, Prafulla Dhariwal, Alex Nichol, Casey Chu, and Mark Chen.
\newblock Hierarchical text-conditional image generation with clip latents.
\newblock \emph{arXiv preprint arXiv:2204.06125}, 1\penalty0 (2):\penalty0 3, 2022.

\bibitem[Retsinas et~al.(2024)Retsinas, Filntisis, Danecek, Abrevaya, Roussos, Bolkart, and Maragos]{retsinas20243d}
George Retsinas, Panagiotis~P Filntisis, Radek Danecek, Victoria~F Abrevaya, Anastasios Roussos, Timo Bolkart, and Petros Maragos.
\newblock 3d facial expressions through analysis-by-neural-synthesis.
\newblock In \emph{Proceedings of the IEEE/CVF Conference on Computer Vision and Pattern Recognition}, pp.\  2490--2501, 2024.

\bibitem[Rombach et~al.(2022)Rombach, Blattmann, Lorenz, Esser, and Ommer]{rombach2022high}
Robin Rombach, Andreas Blattmann, Dominik Lorenz, Patrick Esser, and Bj{\"o}rn Ommer.
\newblock High-resolution image synthesis with latent diffusion models.
\newblock In \emph{Proceedings of the IEEE/CVF conference on computer vision and pattern recognition}, pp.\  10684--10695, 2022.

\bibitem[Ronneberger et~al.(2015)Ronneberger, Fischer, and Brox]{ronneberger2015u}
Olaf Ronneberger, Philipp Fischer, and Thomas Brox.
\newblock U-net: Convolutional networks for biomedical image segmentation.
\newblock In \emph{Medical Image Computing and Computer-Assisted Intervention--MICCAI 2015: 18th International Conference, Munich, Germany, October 5-9, 2015, Proceedings, Part III 18}, pp.\  234--241. Springer, 2015.

\bibitem[Ruan et~al.(2023)Ruan, Ma, Yang, He, Liu, Fu, Yuan, Jin, and Guo]{ruan2023mm}
Ludan Ruan, Yiyang Ma, Huan Yang, Huiguo He, Bei Liu, Jianlong Fu, Nicholas~Jing Yuan, Qin Jin, and Baining Guo.
\newblock Mm-diffusion: Learning multi-modal diffusion models for joint audio and video generation.
\newblock In \emph{Proceedings of the IEEE/CVF Conference on Computer Vision and Pattern Recognition}, pp.\  10219--10228, 2023.

\bibitem[Sagonas et~al.(2016)Sagonas, Antonakos, Tzimiropoulos, Zafeiriou, and Pantic]{sagonas2016300}
Christos Sagonas, Epameinondas Antonakos, Georgios Tzimiropoulos, Stefanos Zafeiriou, and Maja Pantic.
\newblock 300 faces in-the-wild challenge: Database and results.
\newblock \emph{Image and vision computing}, 47:\penalty0 3--18, 2016.

\bibitem[Saharia et~al.(2022)Saharia, Chan, Saxena, Li, Whang, Denton, Ghasemipour, Gontijo~Lopes, Karagol~Ayan, Salimans, et~al.]{saharia2022photorealistic}
Chitwan Saharia, William Chan, Saurabh Saxena, Lala Li, Jay Whang, Emily~L Denton, Kamyar Ghasemipour, Raphael Gontijo~Lopes, Burcu Karagol~Ayan, Tim Salimans, et~al.
\newblock Photorealistic text-to-image diffusion models with deep language understanding.
\newblock \emph{Advances in neural information processing systems}, 35:\penalty0 36479--36494, 2022.

\bibitem[Schuhmann et~al.(2022)Schuhmann, Beaumont, Vencu, Gordon, Wightman, Cherti, Coombes, Katta, Mullis, Wortsman, et~al.]{schuhmann2022laion}
Christoph Schuhmann, Romain Beaumont, Richard Vencu, Cade Gordon, Ross Wightman, Mehdi Cherti, Theo Coombes, Aarush Katta, Clayton Mullis, Mitchell Wortsman, et~al.
\newblock Laion-5b: An open large-scale dataset for training next generation image-text models.
\newblock \emph{Advances in Neural Information Processing Systems}, 35:\penalty0 25278--25294, 2022.

\bibitem[Shen \& Tang(2024)Shen and Tang]{shen2024imagpose}
Fei Shen and Jinhui Tang.
\newblock Imagpose: A unified conditional framework for pose-guided person generation.
\newblock In \emph{The Thirty-eighth Annual Conference on Neural Information Processing Systems}, 2024.

\bibitem[Shen et~al.(2023)Shen, Ye, Zhang, Wang, Han, and Yang]{shen2023advancing}
Fei Shen, Hu~Ye, Jun Zhang, Cong Wang, Xiao Han, and Wei Yang.
\newblock Advancing pose-guided image synthesis with progressive conditional diffusion models.
\newblock \emph{arXiv preprint arXiv:2310.06313}, 2023.

\bibitem[Shen et~al.(2024{\natexlab{a}})Shen, Jiang, He, Ye, Wang, Du, Li, and Tang]{shen2024imagdressing}
Fei Shen, Xin Jiang, Xin He, Hu~Ye, Cong Wang, Xiaoyu Du, Zechao Li, and Jinghui Tang.
\newblock Imagdressing-v1: Customizable virtual dressing.
\newblock \emph{arXiv preprint arXiv:2407.12705}, 2024{\natexlab{a}}.

\bibitem[Shen et~al.(2024{\natexlab{b}})Shen, Ye, Liu, Zhang, Wang, Han, and Yang]{shen2024boosting}
Fei Shen, Hu~Ye, Sibo Liu, Jun Zhang, Cong Wang, Xiao Han, and Wei Yang.
\newblock Boosting consistency in story visualization with rich-contextual conditional diffusion models.
\newblock \emph{arXiv preprint arXiv:2407.02482}, 2024{\natexlab{b}}.

\bibitem[Shen et~al.(2025)Shen, Wang, Gao, Guo, Dang, Tang, and Chua]{shen2025long}
Fei Shen, Cong Wang, Junyao Gao, Qin Guo, Jisheng Dang, Jinhui Tang, and Tat-Seng Chua.
\newblock Long-term talkingface generation via motion-prior conditional diffusion model.
\newblock \emph{arXiv preprint arXiv:2502.09533}, 2025.

\bibitem[Shi et~al.(2024)Shi, Huang, Wang, Bian, Li, Zhang, Zhang, Cheung, See, Qin, et~al.]{shi2024motion}
Xiaoyu Shi, Zhaoyang Huang, Fu-Yun Wang, Weikang Bian, Dasong Li, Yi~Zhang, Manyuan Zhang, Ka~Chun Cheung, Simon See, Hongwei Qin, et~al.
\newblock Motion-i2v: Consistent and controllable image-to-video generation with explicit motion modeling.
\newblock In \emph{ACM SIGGRAPH 2024 Conference Papers}, pp.\  1--11, 2024.

\bibitem[Siarohin et~al.(2019)Siarohin, Lathuili{\`e}re, Tulyakov, Ricci, and Sebe]{siarohin2019first}
Aliaksandr Siarohin, St{\'e}phane Lathuili{\`e}re, Sergey Tulyakov, Elisa Ricci, and Nicu Sebe.
\newblock First order motion model for image animation.
\newblock \emph{Advances in neural information processing systems}, 32, 2019.

\bibitem[Song et~al.(2021{\natexlab{a}})Song, Meng, and Ermon]{song2020denoising}
Jiaming Song, Chenlin Meng, and Stefano Ermon.
\newblock Denoising diffusion implicit models.
\newblock In \emph{ICLR}, 2021{\natexlab{a}}.

\bibitem[Song et~al.(2021{\natexlab{b}})Song, Wu, Fu, Qian, Loy, and He]{song2021everything}
Linsen Song, Wayne Wu, Chaoyou Fu, Chen Qian, Chen~Change Loy, and Ran He.
\newblock Everything's talkin': Pareidolia face reenactment.
\newblock \emph{arXiv preprint arXiv:2104.03061}, 2021{\natexlab{b}}.

\bibitem[Sun et~al.(2013)Sun, Wang, and Tang]{sun2013deep}
Yi~Sun, Xiaogang Wang, and Xiaoou Tang.
\newblock Deep convolutional network cascade for facial point detection.
\newblock In \emph{Proceedings of the IEEE conference on computer vision and pattern recognition}, pp.\  3476--3483, 2013.

\bibitem[Tang et~al.(2023)Tang, Jia, Wang, Phoo, and Hariharan]{tang2023emergent}
Luming Tang, Menglin Jia, Qianqian Wang, Cheng~Perng Phoo, and Bharath Hariharan.
\newblock Emergent correspondence from image diffusion.
\newblock \emph{Advances in Neural Information Processing Systems}, 36:\penalty0 1363--1389, 2023.

\bibitem[Wang et~al.(2024{\natexlab{a}})Wang, Tian, Guan, Zhang, Jiang, Shen, Han, Gu, and Yang]{wang2024ensembling}
Cong Wang, Kuan Tian, Yonghang Guan, Jun Zhang, Zhiwei Jiang, Fei Shen, Xiao Han, Qing Gu, and Wei Yang.
\newblock Ensembling diffusion models via adaptive feature aggregation.
\newblock \emph{arXiv preprint arXiv:2405.17082}, 2024{\natexlab{a}}.

\bibitem[Wang et~al.(2021)Wang, Mallya, and Liu]{wang2021one}
Ting-Chun Wang, Arun Mallya, and Ming-Yu Liu.
\newblock One-shot free-view neural talking-head synthesis for video conferencing.
\newblock In \emph{Proceedings of the IEEE/CVF conference on computer vision and pattern recognition}, pp.\  10039--10049, 2021.

\bibitem[Wang et~al.(2024{\natexlab{b}})Wang, Yuan, Zhang, Chen, Wang, Zhang, Shen, Zhao, and Zhou]{wang2024videocomposer}
Xiang Wang, Hangjie Yuan, Shiwei Zhang, Dayou Chen, Jiuniu Wang, Yingya Zhang, Yujun Shen, Deli Zhao, and Jingren Zhou.
\newblock Videocomposer: Compositional video synthesis with motion controllability.
\newblock \emph{Advances in Neural Information Processing Systems}, 36, 2024{\natexlab{b}}.

\bibitem[Wei et~al.(2024)Wei, Yang, and Wang]{wei2024aniportrait}
Huawei Wei, Zejun Yang, and Zhisheng Wang.
\newblock Aniportrait: Audio-driven synthesis of photorealistic portrait animation.
\newblock \emph{arXiv preprint arXiv:2403.17694}, 2024.

\bibitem[Wu et~al.(2018)Wu, Qian, Yang, Wang, Cai, and Zhou]{wu2018look}
Wayne Wu, Chen Qian, Shuo Yang, Quan Wang, Yici Cai, and Qiang Zhou.
\newblock Look at boundary: A boundary-aware face alignment algorithm.
\newblock In \emph{Proceedings of the IEEE conference on computer vision and pattern recognition}, pp.\  2129--2138, 2018.

\bibitem[Wu et~al.(2017)Wu, Hassner, Kim, Medioni, and Natarajan]{wu2017facial}
Yue Wu, Tal Hassner, KangGeon Kim, Gerard Medioni, and Prem Natarajan.
\newblock Facial landmark detection with tweaked convolutional neural networks.
\newblock \emph{IEEE transactions on pattern analysis and machine intelligence}, 40\penalty0 (12):\penalty0 3067--3074, 2017.

\bibitem[Xie et~al.(2024)Xie, Xu, Song, Wang, Shi, and Luo]{xie2024x}
You Xie, Hongyi Xu, Guoxian Song, Chao Wang, Yichun Shi, and Linjie Luo.
\newblock X-portrait: Expressive portrait animation with hierarchical motion attention.
\newblock In \emph{ACM SIGGRAPH 2024 Conference Papers}, pp.\  1--11, 2024.

\bibitem[Xing et~al.(2023)Xing, Xia, Zhang, Chen, Wang, Wong, and Shan]{xing2023dynamicrafter}
Jinbo Xing, Menghan Xia, Yong Zhang, Haoxin Chen, Xintao Wang, Tien-Tsin Wong, and Ying Shan.
\newblock Dynamicrafter: Animating open-domain images with video diffusion priors.
\newblock \emph{arXiv preprint arXiv:2310.12190}, 2023.

\bibitem[Xu et~al.(2022)Xu, Jin, Zeng, Liu, Qian, Ouyang, Luo, and Wang]{xu2022pose}
Lumin Xu, Sheng Jin, Wang Zeng, Wentao Liu, Chen Qian, Wanli Ouyang, Ping Luo, and Xiaogang Wang.
\newblock Pose for everything: Towards category-agnostic pose estimation.
\newblock In \emph{European conference on computer vision}, pp.\  398--416. Springer, 2022.

\bibitem[Yang et~al.(2023)Yang, Zeng, Zhang, and Zhang]{yang2023unipose}
Jie Yang, Ailing Zeng, Ruimao Zhang, and Lei Zhang.
\newblock Unipose: Detecting any keypoints.
\newblock \emph{arXiv preprint arXiv:2310.08530}, 2023.

\bibitem[Yang et~al.(2024)Yang, Li, Wu, Jing, Li, Ji, Liang, and Fan]{yang2024megactor}
Shurong Yang, Huadong Li, Juhao Wu, Minhao Jing, Linze Li, Renhe Ji, Jiajun Liang, and Haoqiang Fan.
\newblock Megactor: Harness the power of raw video for vivid portrait animation.
\newblock \emph{arXiv preprint arXiv:2405.20851}, 2024.

\bibitem[Ye et~al.(2023)Ye, Zhang, Liu, Han, and Yang]{ye2023ip}
Hu~Ye, Jun Zhang, Sibo Liu, Xiao Han, and Wei Yang.
\newblock Ip-adapter: Text compatible image prompt adapter for text-to-image diffusion models.
\newblock \emph{arXiv preprint arXiv:2308.06721}, 2023.

\bibitem[Zeng et~al.(2023)Zeng, Liu, Gao, Liu, Li, Liu, and Zhang]{zeng2023face}
Bohan Zeng, Xuhui Liu, Sicheng Gao, Boyu Liu, Hong Li, Jianzhuang Liu, and Baochang Zhang.
\newblock Face animation with an attribute-guided diffusion model.
\newblock In \emph{Proceedings of the IEEE/CVF Conference on Computer Vision and Pattern Recognition}, pp.\  628--637, 2023.

\bibitem[Zhang et~al.(2020)Zhang, Zeng, Wang, Pan, Liu, Liu, Ding, and Fan]{zhang2020freenet}
Jiangning Zhang, Xianfang Zeng, Mengmeng Wang, Yusu Pan, Liang Liu, Yong Liu, Yu~Ding, and Changjie Fan.
\newblock Freenet: Multi-identity face reenactment.
\newblock In \emph{Proceedings of the IEEE/CVF conference on computer vision and pattern recognition}, pp.\  5326--5335, 2020.

\bibitem[Zhang et~al.(2023{\natexlab{a}})Zhang, Wang, Zhang, Zhao, Yuan, Qin, Wang, Zhao, and Zhou]{zhang2023i2vgen}
Shiwei Zhang, Jiayu Wang, Yingya Zhang, Kang Zhao, Hangjie Yuan, Zhiwu Qin, Xiang Wang, Deli Zhao, and Jingren Zhou.
\newblock I2vgen-xl: High-quality image-to-video synthesis via cascaded diffusion models.
\newblock \emph{arXiv preprint arXiv:2311.04145}, 2023{\natexlab{a}}.

\bibitem[Zhang et~al.(2023{\natexlab{b}})Zhang, Zhai, Wei, Yang, and Ma]{zhang2023blind}
Weixia Zhang, Guangtao Zhai, Ying Wei, Xiaokang Yang, and Kede Ma.
\newblock Blind image quality assessment via vision-language correspondence: A multitask learning perspective.
\newblock In \emph{Proceedings of the IEEE/CVF conference on computer vision and pattern recognition}, pp.\  14071--14081, 2023{\natexlab{b}}.

\bibitem[Zhang et~al.(2024)Zhang, Xing, Zeng, Fang, and Chen]{zhang2024pia}
Yiming Zhang, Zhening Xing, Yanhong Zeng, Youqing Fang, and Kai Chen.
\newblock Pia: Your personalized image animator via plug-and-play modules in text-to-image models.
\newblock In \emph{Proceedings of the IEEE/CVF Conference on Computer Vision and Pattern Recognition}, pp.\  7747--7756, 2024.

\bibitem[Zhang et~al.(2021)Zhang, Li, Ding, and Fan]{zhang2021flow}
Zhimeng Zhang, Lincheng Li, Yu~Ding, and Changjie Fan.
\newblock Flow-guided one-shot talking face generation with a high-resolution audio-visual dataset.
\newblock In \emph{Proceedings of the IEEE/CVF Conference on Computer Vision and Pattern Recognition}, pp.\  3661--3670, 2021.

\bibitem[Zhao \& Zhang(2022)Zhao and Zhang]{zhao2022thin}
Jian Zhao and Hui Zhang.
\newblock Thin-plate spline motion model for image animation.
\newblock In \emph{Proceedings of the IEEE/CVF Conference on Computer Vision and Pattern Recognition}, pp.\  3657--3666, 2022.

\bibitem[Zhou et~al.(2013)Zhou, Fan, Cao, Jiang, and Yin]{zhou2013extensive}
Erjin Zhou, Haoqiang Fan, Zhimin Cao, Yuning Jiang, and Qi~Yin.
\newblock Extensive facial landmark localization with coarse-to-fine convolutional network cascade.
\newblock In \emph{Proceedings of the IEEE international conference on computer vision workshops}, pp.\  386--391, 2013.

\bibitem[Zhou et~al.(2023)Zhou, Li, Liu, Wang, Yu, and Ji]{zhou2023star}
Zhenglin Zhou, Huaxia Li, Hong Liu, Nanyang Wang, Gang Yu, and Rongrong Ji.
\newblock Star loss: Reducing semantic ambiguity in facial landmark detection.
\newblock In \emph{Proceedings of the IEEE/CVF conference on computer vision and pattern recognition}, pp.\  15475--15484, 2023.

\end{thebibliography}
\bibliographystyle{iclr2025_conference}

\newpage
\appendix
\section*{Appendix}

\section{Implementation Details}
\label{sec:app_imp}
In FaceShot, we use a single H800 to generate animation results.
And we have included a total of 46 images and 24 driving videos in CharacBench, with each video consisting of 110 to 127 frames.
All videos (.mp4) and images (.jpg) are processed into a resolution of $512\times 512$.
Following MOFA-Video, for human face and driving video, we utilized Facial Alignment Network (FAN) implemented in facexlib as our annotating algorithm to detect the landmarks.
And for non-human characters, we perform manual annotating as even the SOTA face landmark detection methods still fail on these characters.

\section{Methods}
\noindent{\textbf{Appearance Matching in Appearance Gallery.}} 
As mentioned in Section \ref{sec:method}, the purpose of the appearance gallery is to reduce appearance discrepancies by matching the reference image to the closest target domain. 
In detail, the reference image is first cropped into five facial parts i.e., eyes, mouth, nose, eyebrows and face boundary as:
\begin{equation*}
I_{ref} = [I_{ref, e}, I_{ref, m}, I_{ref, n}, I_{ref, eb}, I_{ref, fb}],
\end{equation*}
each facial part includes a specific number of landmarks, as listed in the Table \ref{tab:land_num}:

\begin{table}[h]
    \centering
    \caption{Specific landmark number for each facial part.}
    \begin{tabular}{ccccc}
    \toprule
       eyes & mouth & nose & eyebrows & face boundary \\
         \hline
        12 & 20 & 9 & 10 & 17 \\
        \bottomrule
    \end{tabular}
    \label{tab:land_num}
\end{table}
Next, each part is matched to the closest target domain in the appearance gallery by calculating the average CLIP image score:
\begin{equation*}
 G^*_p  = \underset{j \in D}{\arg \max}\ \ \text{CLIP-S}(I_{ref,p}, \frac{1}{k}\sum^k_{i=1} I_{tar,p}^{j,i}),
\end{equation*}
where $p\in \{e, m, n, eb, fb\}$ and $D$ represents the domains in part $p$, $k$ denotes the number of target images in given domain and CLIP-S denotes the clip image score.
Finally, the target images are formulated as $I_{tar} = [G_e^*, G_m^*, G_n^*, G_{eb}^*, G_{fb}^*]$.

\begin{wrapfigure}{r}{0.5\linewidth}
\centering
    \vspace{-4mm}
    \includegraphics[width=0.85\linewidth]{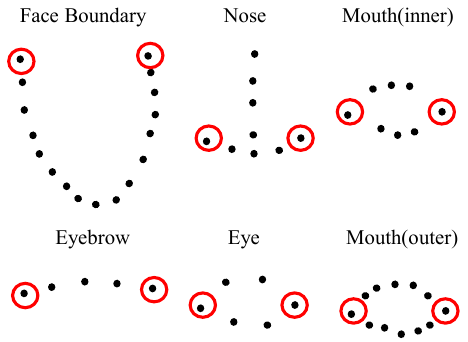}
    \caption{Illustration of the endpoints in each part, marked with a red circle.}
    \vspace{-6mm}
    \label{fig:endpoints}
\end{wrapfigure}

\begin{figure*}[t!]
    \centering
        \vspace{-4mm}
    \includegraphics[width=\linewidth]{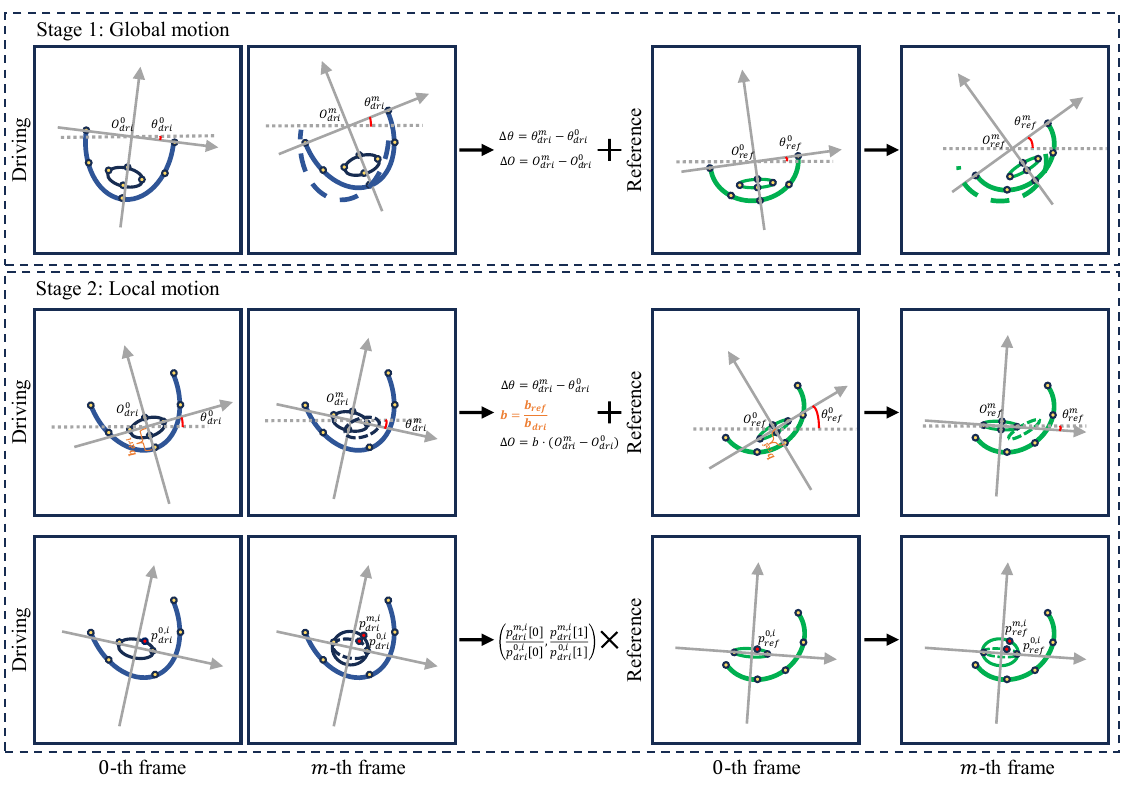}
    \caption{Illustration of our coordinate-based landmark retargeting module. Specifically, our module consists of two stages: global motion and local motion retargeting, which aim to capture global and local positional changes of the entire face and individual facial parts separately.}
    \vspace{-4mm}
    \label{fig:illustration}
\end{figure*}

\noindent{\textbf{Settings in Landmark Retargeting.}}
The angle and the origin of the rectangular coordinate system are calculated using two endpoints $p^{e_1}$ and $p^{e_2}$ as follows:
\begin{equation*}
\begin{aligned}
       & O =\left(\frac{p^{e_1}[0]+p^{e_2}[0]}{2}, \frac{p^{e_1}[1]+p^{e_2}[1]}{2}\right), \\
       & \theta =\text{arctan}\left(\frac{p^{e_2}[1]-p^{e_1}[1]}{p^{e_2}[0]-p^{e_1}[0]}\right).
\end{aligned}
\end{equation*}
And the indices of the endpoints within each part of every frame are fixed, as illustrated in the Figure \ref{fig:endpoints}.  
For better understanding of our coordinate-based landmark retargeting module, we provide an illustration of this module in Figure \ref{fig:illustration}.

\section{Experiments}
\noindent{\textbf{Choices of Time Step $t$ and Layer $l$.}} 
To extract the diffusion feature that best fits facial instances, we conduct detailed experiments on the selection of time steps $t$ and layer $l$ of U-Net. 
Specifically, we test different combinations of $t$ and $l$ on 300W \citep{sagonas2016300}, a widely used facial dataset, and report NME as the quantitative result.
As shown in Figure \ref{fig:heatmap}, we achieve the best NME value when $t=301$ and $l=6$, which are used as the basic settings of our paper.

\begin{wrapfigure}{r}{0.4\linewidth}
\centering
\vspace{-4mm}
    \includegraphics[width=1.0\linewidth]{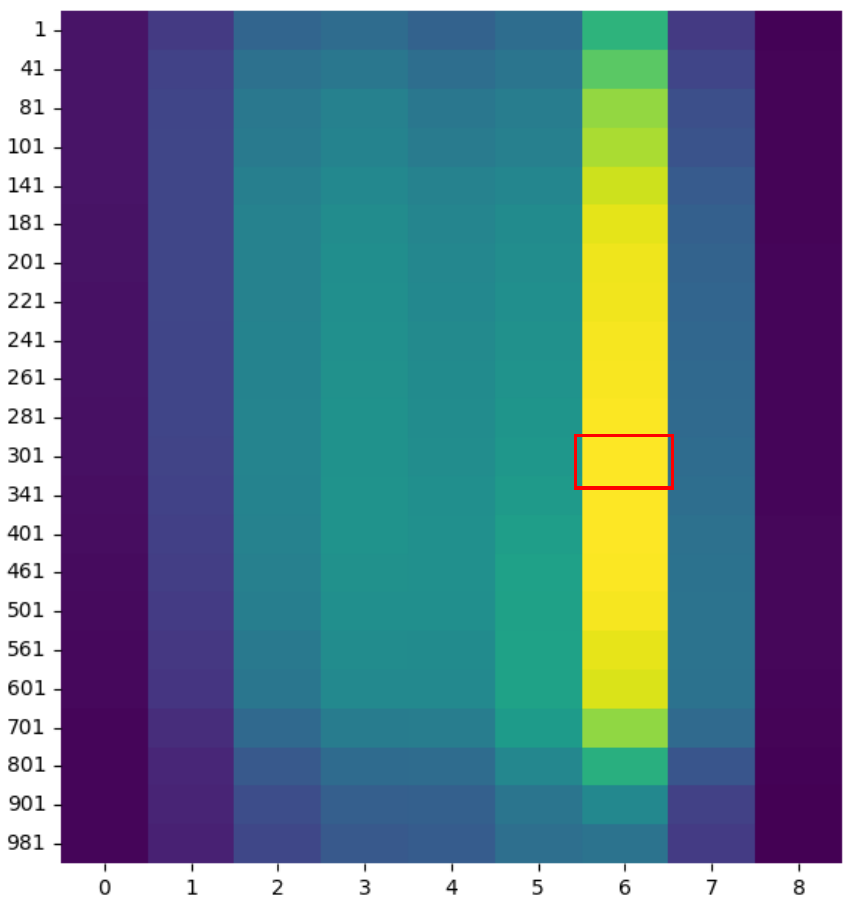}
    \caption{Heatmap of NME values for time step $t$ and layer $l$ of the U-Net.}
    \vspace{-4mm}
    \label{fig:heatmap}
\end{wrapfigure}
\noindent{\textbf{Choice of Target Number $k$.}} 
As mentioned in Section \ref{sec:method}, we use the averaged diffusion feature at the $i$-th point of $k$ target images to improve matching performance. We evaluate different values of $k = 1, 5, 10, 15, 50, 1000$ on the 300W dataset and report the NME in Table~\ref{tab:k}. Results indicate that increasing $k$ significantly improves performance when $k \leq 10$. However, for $k > 10$, the time cost increases exponentially with diminishing performance gains. Based on this observation, we set $k=10$ for our experiments.
Please note that this analysis is solely aimed at determining the number of target images. However, the features of the target images will be pre-stored as local data, which is not considered additional time overhead during inference.

\begin{figure*}
\centering
\begin{minipage}{0.54\linewidth}
\centering
    \includegraphics[width=0.94\linewidth]{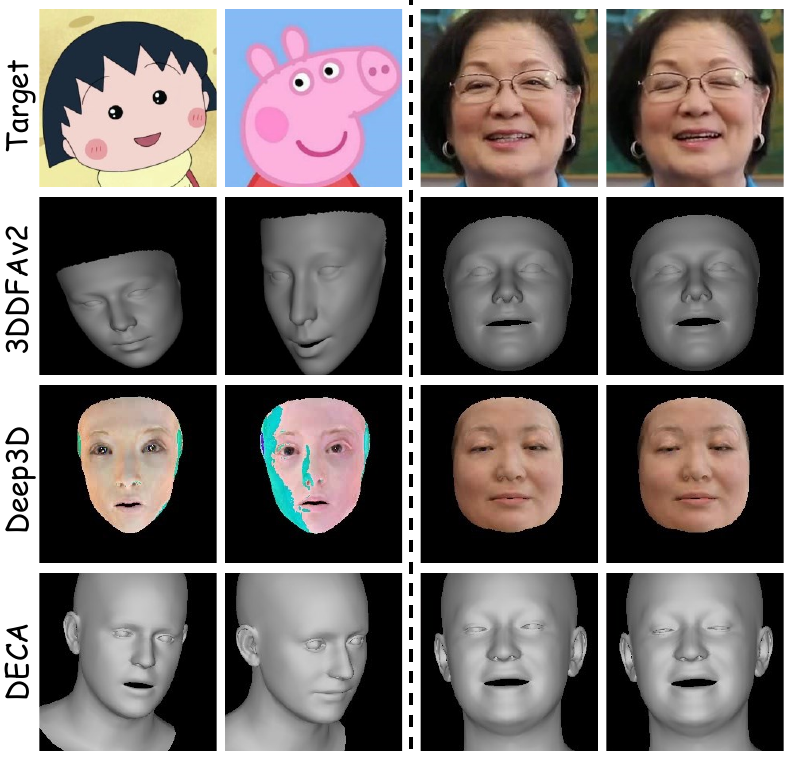}
    \captionof{figure}{Modeling results of different 3DMM methods.}
    \label{fig:more_3d}
\end{minipage}
\begin{minipage}{0.4\linewidth}
\setcounter{figure}{16}
\begin{minipage}{\linewidth}
\centering
    \includegraphics[width=0.98\linewidth]{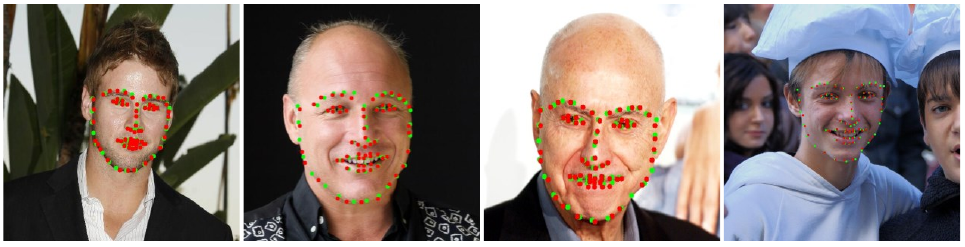}
    \captionof{figure}{Visual results of landmarks on human faces.}
    \label{fig:300w}
\end{minipage}
\\
\centering
\begin{minipage}{\linewidth}
    \includegraphics[width=\linewidth]{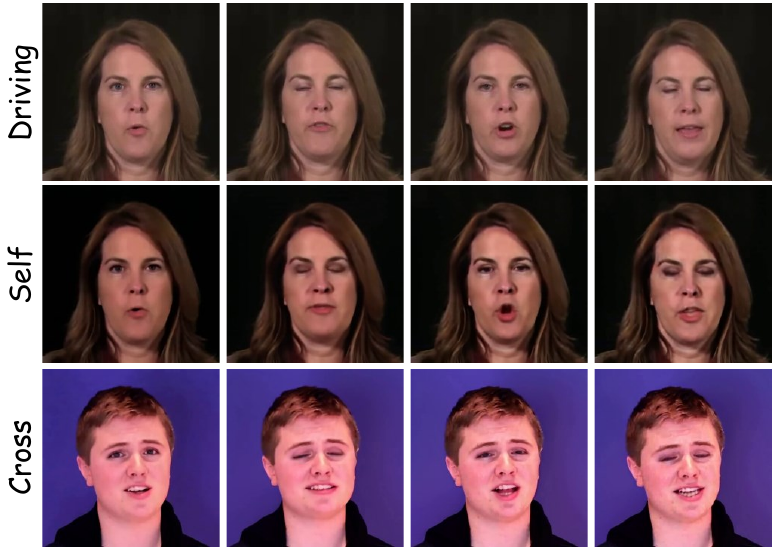}
    \captionof{figure}{Self and cross identity driving results on HDTF.}
    \label{fig:hdtf}
\end{minipage}
\end{minipage}

\vspace{-4mm}
\end{figure*}

\noindent{\textbf{3DMM Modeling.}}
Following our base model MOFA-Video, we adopt Deep3D \citep{deng2019accurate} as the 3DMM method in Figure \ref{fig:3dmm}. Deep3D employs a deep network to predict 3D coefficients (coeff) at each frame of driven videos, instead of iterative fitting.
Additionally, we provide the 3D modeling results of DECA \citep{feng2021learning}, Deep3D and 3DDFAv2 \citep{guo2020towards} on non-human characters and driven videos in Figure \ref{fig:more_3d}. Our observations reveal that none of these 3DMM methods can accurately generate precise 3D models of non-human characters or capture subtle movements in driven sequences, such as eye closure.

\noindent{\textbf{Visualization of Appearance Guidance.}}
We provide the cosine similarity distribution during inference. 
As depicted in Figure \ref{fig:distribution}, with prior appearance knowledge, the similarity between reference points and unrelated target points becomes smaller, reducing the probability of mismatching.

\setcounter{figure}{15}
\begin{wrapfigure}{r}{0.4\linewidth}
\centering
    \includegraphics[width=0.8\linewidth]{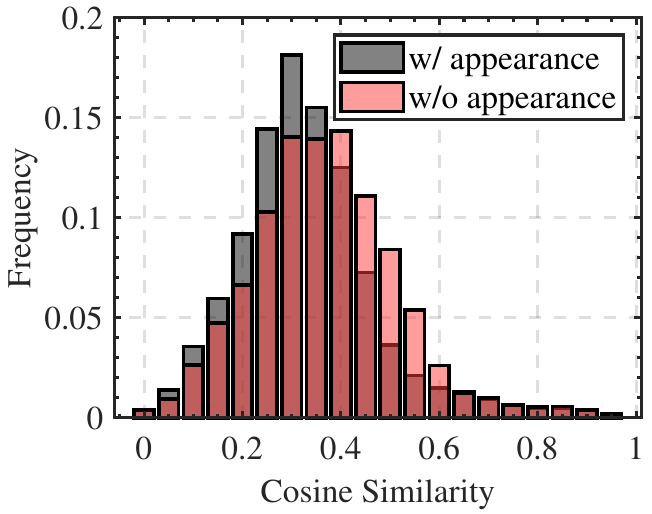}
    \vspace{-4mm}
    \caption{Cosine similarity distribution with or without appearance guidance.}
    \vspace{-6mm}
    \label{fig:distribution}
\end{wrapfigure}
\noindent{\textbf{Human Evaluations.}}
To evaluate the effectiveness of FaceShot on human faces, we first provide the visual landmark results on 300W in Figure \ref{fig:300w}.

Furthermore, we present the animation results for self-identity and cross-identity driving on traditional facial video datasets HDTF \citep{zhang2021flow} in Figure \ref{fig:hdtf}. FaceShot perform well on these real human faces, showing its robustness.

\noindent{\textbf{Time Efficiency.}}
We present a time analysis of each step in FaceShot for processing varying numbers of frames on a single H800 GPU, as shown in the Table \ref{tab:time}:
\begin{itemize}
\item \textit{Driving Detection}: Detecting the landmark sequence of the driving video using the landmark detector from MOFA-Video. The detector used is FAN from facexlib.
\item \textit{Landmark Matching}: Detecting the target image landmarks using the appearance-guided landmark matching module. The time cost of landmark matching remains almost identical regardless of the number of frames because the matching is required only once for the target image, irrespective of the number of frames in the driving video. The 0.8-second time includes both the DDIM inversion and the argmin operation in Eq. \ref{eq:opt}.
\item \textit{Landmark Retargeting}: Retargeting landmark motion using coordinate-based landmark retargeting module. As no model parameters are required, our retargeting modules can generate precise landmark sequences with very low time cost.
\end{itemize}

\begin{table}[h]
\centering
\vspace{-4mm}
    \caption{Experiments on the number $k$ of target images. Best NME result is marked in \textbf{bold}.}
    \vspace{-2mm}
    \begin{tabular}{c|cccccc}
    \toprule
     Metric &1 & 5 &10& 15 &50 &1000\\
     \hline
    NME$\downarrow$ & 12.801 &7.009& 6.343&6.267& 6.252&\textbf{6.104}\\
    \bottomrule
    \end{tabular}
    \vspace{-2mm}
    \label{tab:k}
\end{table}
\begin{table}[h]
    \centering
    \vspace{-2mm}
        \caption{Time analysis of each step in FaceShot for processing varying numbers of frames.}
        \vspace{-2mm}
    \begin{tabular}{cccc}
    \toprule
       frames & \makecell{Driving \\ Detection} & \makecell{Landmark\\ Matching} & \makecell{Landmark\\ Retargeting} \\
         \hline
        50 & 1.817(s) & 0.860(s) & 0.382(s) \\
        100 & 3.562(s) & 0.858(s) & 0.751(s) \\
        \bottomrule
    \end{tabular}
    \label{tab:time}
\end{table}

In conclusion, FaceShot introduces only a 119ms additional time overhead when used as a plugin for MOFA-Video (for 50 frames). 
This minimal time cost is negligible compared to the inference time of diffusion-based models (approximately 80 seconds for 50 frames).
As shown in Table \ref{tab:inf_time}, FaceShot achieves the low time cost among diffusion-based methods, including AniPortrait, FADM, Follow Your Emoji, MegActor, X-Portrait, and MOFA-Video.

\begin{table}[h]
    \centering
    \setlength\tabcolsep{2pt} 
        \caption{Time analysis SOTA methods inference on 50 frames. Symbol $^*$ represents GAN-based method.}
    \begin{tabular}{cccc}
    \toprule
       Methods & Time & Methods & Time \\
         \hline
        $\text{FaceVid2Vid}^*$ & 4.308(s) & MegActor & 174.189(s) \\
        $\text{LivePortrait}^*$ & 4.321(s) & X-Portrait & 132.702(s) \\
        AniPortrait & 99.977(s) & MOFA-Video & 79.421(s) \\
        FADM & 88.368(s) & FaceShot & 79.540(s)
 \\
        \makecell{Follow Your Emoji} & 112.830(s) &  &   \\
        \bottomrule
    \end{tabular}

    \label{tab:inf_time}
\end{table}
\end{document}